\documentclass[11pt]{article}

\usepackage[preprint]{acl}

\usepackage{times}
\usepackage{latexsym}

\usepackage[T1]{fontenc}

\usepackage[utf8]{inputenc}

\usepackage{microtype}

\usepackage{inconsolata}

\usepackage{graphicx}

\usepackage{algorithm}
\usepackage{algorithmic}

\usepackage{amsmath}
\usepackage{amsfonts}

\usepackage{tikz}
\usepackage{pgfplots}
\pgfplotsset{compat=1.18}

\usepackage{placeins}

\title{MD-ProTector: Positioning Multiple Data-Driven Prototypes for LLM-Generated Text Detection}

\author{
Jinmo Han \quad Jimin Hong \quad Chanyeong Moon \\
\textbf{Ju Yeon Kang \quad Seonuk Kim \quad Nam Soo Kim} \\[2pt]
Department of Electrical and Computer Engineering and INMC \\
Seoul National University, Seoul, Republic of Korea \\
\texttt{\{jinmo, jimin, chanyeong\}@hi.snu.ac.kr} \\
\texttt{\{juyeon, seonuk\}@hi.snu.ac.kr}
\quad
\texttt{nkim@snu.ac.kr}
}

\begin{document}
\maketitle

\begin{abstract}
As LLM-generated content becomes more sophisticated, detection systems for distinguishing those texts from human-written text must operate at scale while handling diverse writing styles, domains, languages, and generator models. 
Input-only encoder detectors are suitable for practical deployment setting, but standard binary classification supplies only the class label and does not explicitly organize the substantial variation within either class.
We propose \textbf{MD-ProTector}, which represents each class with multiple trainable reference vectors in the encoder embedding space, referred to as prototypes.
These prototypes provide separate decision boundaries for different groups of texts within the same class.
However, adding multiple prototypes alone does not determine which variation each prototype should represent.
MD-ProTector addresses this problem with Prototype Positioning loss, which separates class-level structure from the within-class variation that differentiates individual prototypes.
Evaluated across five settings from three large-scale benchmarks covering domain, generator, language, and adversarial variation, MD-ProTector achieves the highest AvgRec on MAGE CDCM and RAID and the highest AUROC and lowest FPR95 on RAID among the compared encoder-based methods.\footnote{Code and configurations: \url{https://github.com/WiFiHan/md-protector}}

\end{abstract}

\section{Introduction}

As large language models have become increasingly sophisticated and widely accessible, detecting whether text is human-written or LLM-generated has become essential for ensuring credibility in digital communication \citep{kwon_2025_survey}. 
This is particularly important for blocking automated fake news and phishing attacks, as well as maintaining academic integrity \citep{najjar2025detecting}. 
In large-scale deployment scenarios, detection systems must be applied to massive volumes of content that span diverse writing styles, domains, and generator models \citep{li-etal-2024-mage}. 
Treating such diversity is important for practical detection systems operating in real-world cases including various domain, writing style, generator, and adversarial editing cases \citep{wu2024detectrl}.

\begin{figure}[t]
    \centering
    \includegraphics[width=\linewidth]{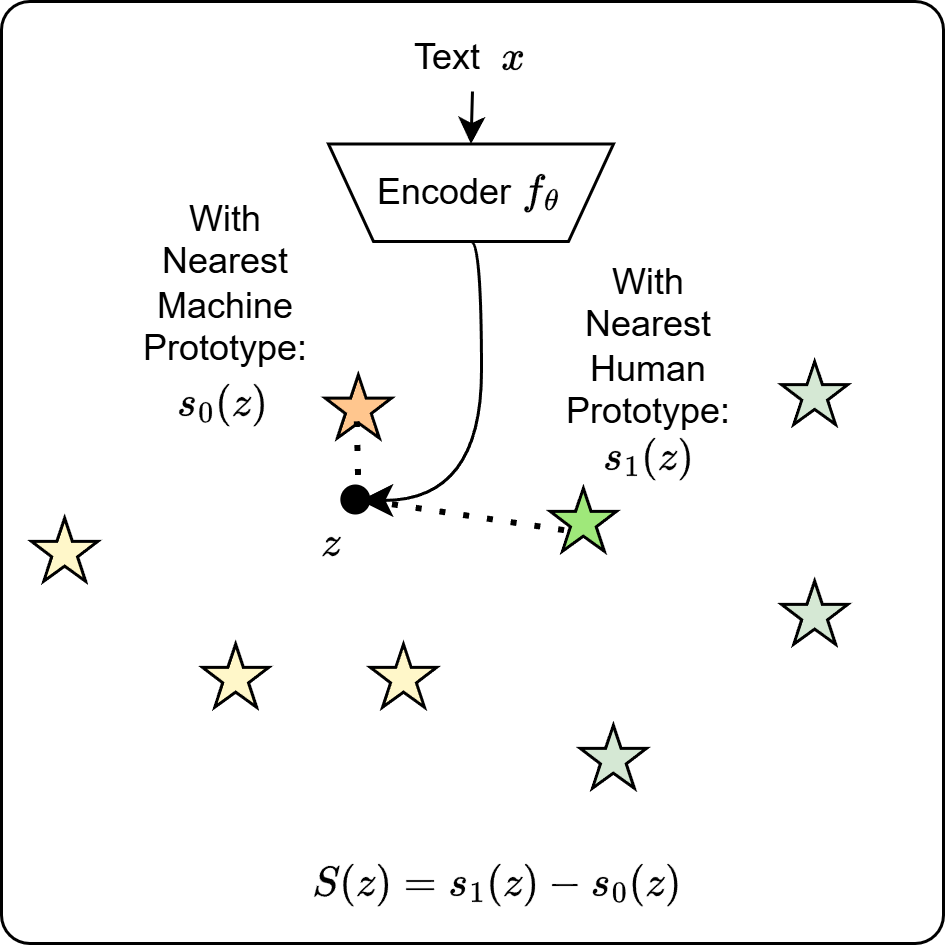}
    \caption{
    \textbf{Prototype-Based Inference.}
    Given an input text $x$, MD-ProTector encodes it into a normalized embedding $z$ and compares it with the machine and human prototype banks.
    The detection score is $S(z)=s_1(z)-s_0(z)$, where $s_0(z)$ and $s_1(z)$ are the maximum similarities to the machine and human prototypes, respectively.
    }
    \label{fig:inference}
\end{figure}

In these practical deployment settings, watermarks or model-internal scores such as log-likelihoods are not guaranteed to be available \citep{christ2024undetectable, mitchell2023detectgpt}.
Under these conditions, detectors based on lightweight text encoders provide a practical solution. 
They operate directly on the input text without requiring access to the generation pipeline or model internals \citep{kuznetsov-etal-2024-robust}. 
This also allows one encoder pipeline to be adapted across target domains without access to the generator internals \citep{rodriguez-etal-2022-cross}.

One of the simplest designs for encoder-based detectors is to attach a binary classification head and train it using a cross-entropy loss \citep{ippolito-etal-2020-automatic}. 
Despite such intra-class diversity in writing style and domain, all texts are reduced to two categories during training under this design.
The standard binary classification objective is a coarse supervision signal that does not explicitly consider such intra-class diversity \citep{cui2016fine}.
Several prior works have attempted to move beyond standard binary classification formulations by adding structural constraints, but still overlook intra-class diversity within at least one of the two classes, which can limit detection performance \citep{guo2024detective, zeng2025human-ood}. 
This observation motivates detectors that represent both human-written and LLM-generated texts with multiple local representatives rather than a single global class representation.

A natural extension is to represent each class with multiple trainable reference vectors in the embedding space, which we call prototypes.
However, the number of prototypes alone does not determine which pattern of within-class variation each prototype should represent.
Without a prototype-specific objective, different prototypes may remain redundant or capture overlapping patterns.

We therefore propose \textbf{MD-ProTector}, an input-only encoder detector that learns data-driven positions for separate human and machine prototypes.
MD-ProTector separates the direction shared within each class from the variation that distinguishes groups of samples within that class.
Prototype Positioning loss uses this variation to give different prototypes distinct roles, while complementary objectives keep each prototype aligned with the corresponding human or machine class.
The learned prototype banks then directly define the human--machine detection score shown in Figure~\ref{fig:inference}.

We evaluate MD-ProTector under mixed domain and generator conditions, adversarial perturbations, multilingual generalization, held-out domains and held-out generator models.
Among input-only encoder detectors trained under the same data, backbone, and validation protocol, MD-ProTector ranks within the top two in AvgRec across all five settings.
It achieves the highest AvgRec on MAGE CDCM and RAID, together with the highest AUROC and lowest FPR95 on RAID, while DeTeCtive remains stronger on M4.
Ablation studies show Prototype Positioning performs better than other prototype multiplicity methods such as direct prototype repulsion and positioning before removal of the class-shared direction.

\medskip
Our contributions are as follows:
\begin{itemize}
    \item We develop an input-only LLM-generated text detector in which separate human and machine prototype banks directly define the detection score.
    \item We introduce the Prototype Positioning loss to place each prototype to capture within-class variation in data-driven manner.
    \item We evaluate MD-ProTector across five controlled input-only encoder settings, where it leads AvgRec on MAGE and RAID and attains the highest AUROC and lowest FPR95 on RAID.
\end{itemize}
\section{Related Works}

\subsection{LLM-Generated Text Detection}
\label{subsec:llm-detection}

Prior work on LLM-generated text detection includes watermarking, zero-shot statistical methods, and supervised classifiers, depending on model access and the types of detection signals used \citep{wu-etal-2025-survey}. 
Watermarking injects identifiable patterns during generation, enabling content source verification by model providers \citep{pmlr-v202-kirchenbauer23a}. 
Zero-shot statistical detectors, such as GLTR, DetectGPT, Fast-DetectGPT, and Binoculars, use token probabilities, probability curvature, or cross-model likelihood ratios without task-specific training \citep{gehrmann-etal-2019-gltr,mitchell2023detectgpt,bao2024fastdetectgpt,hans2024binoculars}. 
BISCOPE, ImBD, DetectAnyLLM, and PAWN further refine model-based scoring through memorization, style-aligned discrepancy, task-oriented discrepancy learning, or learned token weighting \citep{guo2024biscope,chen2025imbd,fu2025detectanyllm,miralles2025pawn}. 
Rewrite-based methods instead use the amount or pattern of changes produced by an auxiliary LLM \citep{mao2024raidar,hao2025learning2rewrite}. 
These methods can provide strong detection signals, but require access to scoring language models, auxiliary generation calls, or cooperation from the generator.

As a practical alternative in large-scale deployment scenarios, supervised detectors can operate directly on input text and are suitable for in-the-wild deployment \citep{bakhtin-etal-2019-real,uchendu-etal-2020-authorship,wang-etal-2023-implementing}. 
These detectors typically rely on encoder architectures such as BERT \citep{devlin-etal-2019-bert} or RoBERTa \citep{liu2019roberta}. 
Ghostbuster builds a classifier from features extracted by several weaker language models, RADAR improves paraphrase robustness through adversarial training, and MoSEs models stylistic references with input-dependent threshold estimation \citep{verma2024ghostbuster,hu-etal-2023-radar,wu2025moses}. 
Despite their efficiency, supervised detectors can still degrade when the domain or generator distribution changes \citep{bakhtin-etal-2019-real,li-etal-2024-mage,wu2024detectrl}.

Recent work therefore imposes stronger structure on the representation space.
DeTeCtive organizes text instances through author- and style-aware contrastive supervision and performs KNN inference \citep{guo2024detective}.
DSVDD takes a one-class approach, compacting machine-generated embeddings and treating human-written text as out-of-distribution \citep{zeng2025human-ood}.
SAMP represents both classes with multiple prototypes using source-model supervision \citep{xu2026samp}.
While these methods introduce instance-level structure, one-class compactness, or source-aware prototypes, binary supervision alone does not specify how internal variation should form distinct training targets for multiple human and machine prototypes.
MD-ProTector addresses this gap by constructing a separate positioning target for each prototype from the hub-removed residuals of its associated samples.

\subsection{Prototype-Based Representation Learning}
\label{subsec:prototype-based}

Prototype-based methods represent each class by one or more representative points in an embedding space and classify inputs based on their similarity to these prototypes.
A canonical example is Prototypical Networks, which compute class prototypes as the mean embeddings of support examples and classify queries by distance in a metric-learning framework \citep{snell2017prototypical}.
ProtoFewRoBERTa applies this episodic formulation to few-shot detection of AI-generated reviews, while ProtoryNet learns sentence-level reference patterns and classifies documents from their prototype trajectories \citep{agrahari2025protofew,hong2023protorynet}.
These methods establish prototypes as data-driven class summaries or interpretable reference patterns.

Prototype-based representations have also been applied beyond standard classification.
Prototypical Contrastive Learning estimates prototypes as latent cluster variables for representation learning \citep{liprototypical}.
OOD methods use class prototypes, diversified prototypes, or mixtures of prototypes to represent the known data distribution and score unfamiliar inputs \citep{chen2024proto,jia2025diversified,lu2024palm}.
Multiple normal prototypes have similarly been used in anomaly detection, and multi-prototype modeling has been applied to open-set noisy-label learning \citep{dong2024multi,zhang2025multiprototype}.

Prior multi-prototype methods learn prototypes from full sample embeddings or through assignment and separation objectives, leaving the respective roles of the class-shared direction and the variation that differentiates individual prototypes underdetermined.
Without this distinction, the learning objective does not specify which component should preserve the class decision and which component should organize multiple representatives within the class.
MD-ProTector resolves this ambiguity by preserving the shared direction through the class hub and using assignment-weighted, hub-orthogonal residuals to position each prototype.
\section{Proposed Method}
\label{sec:method}

\begin{figure*}[t]
\centering
\includegraphics[width=\textwidth]{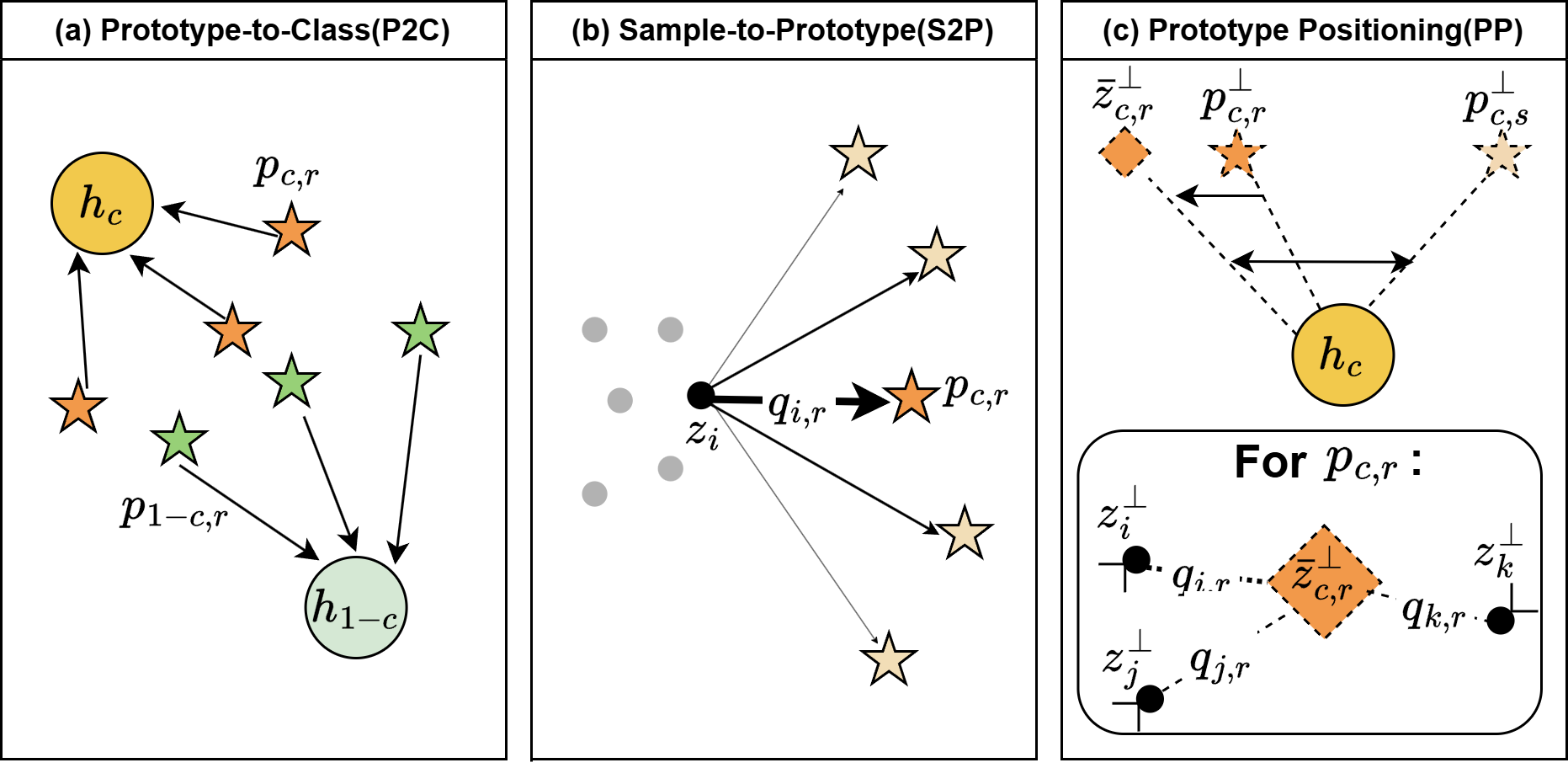}
\caption{
Overview of the training objectives.
Circles denote class hubs, stars denote learnable prototypes, dots denote sample embeddings, and diamonds denote assignment-weighted aggregates of sample residuals.
Prototype-to-Class aligns prototypes with their class hubs.
Sample-to-Prototype associates samples with the prototype bank of their ground-truth class.
Prototype Positioning aligns the residual vector of each prototype with the aggregate constructed for that prototype after removing the class-hub direction.
In panel (c), $p_{c,s}^{\perp}$ denotes the residual vector of another prototype with $s\neq r$.
}
\label{fig:training_loss}
\end{figure*}

MD-ProTector jointly learns an encoder and separate prototype banks for human-written and LLM-generated text.
Prototype Positioning organizes each prototype according to the residual variation of its associated samples, while Prototype-to-Class and Sample-to-Prototype preserve class alignment and sample association.
The learned prototype banks are used directly for detection.

\subsection{Problem Formulation and Encoder}

Given an input text $x_i$, each sample has a binary label $y_i\in\{0,1\}$, where $y_i=0$ denotes LLM-generated text and $y_i=1$ denotes human-written text.
A lightweight encoder $f_\theta$ maps the input into token-level representations, which are mean-pooled and normalized as
\begin{equation}
    z_i
    =
    \mathrm{norm}\left(f_\theta(x_i)\right),
    \label{eq:sample-embedding}
\end{equation}
where
$\mathrm{norm}(v)
=
v/\lVert v\rVert_2$
denotes $\ell_2$ normalization.

For a mini-batch $B$ in which both classes are represented, let
$B_c=\{i\in B:y_i=c\}$.
The class hub is
\begin{equation}
    h_c
    =
    \mathrm{norm}
    \left(
    \frac{1}{|B_c|}
    \sum_{i\in B_c} z_i
    \right).
    \label{eq:class-hub}
\end{equation}
It represents the direction shared by class-$c$ samples in the current mini-batch.

For each class $c\in\{0,1\}$, we maintain $R$ learnable prototypes:
\begin{equation}
    \mathcal P_c
    =
    \{p_{c,1},\ldots,p_{c,R}\},
    \qquad
    \lVert p_{c,r}\rVert_2=1.
    \label{eq:prototype-bank}
\end{equation}
The full prototype bank is
$\mathcal P=\mathcal P_0\cup\mathcal P_1$.
The objectives below preserve class-level alignment while allowing different groups of same-class samples to orient the residual component of each prototype.

\subsection{Data-Driven Prototype Initialization}

Before training, we extract embeddings from the training set and apply K-Means separately within each class \citep{mcqueen1967some}.
The resulting centroids initialize
$\mathcal P_c^{(0)}
=
\{p_{c,r}^{(0)}\}_{r=1}^{R}$.
The centroids are normalized and subsequently optimized as learnable parameters together with the encoder.
This initialization places the prototype banks in the observed class distributions rather than at random directions.

\subsection{Training Objectives}

For compact notation, define
\begin{equation}
    \phi(u,v)
    =
    \exp\!\left(u^\top v/\tau\right),
    \label{eq:phi}
\end{equation}
where $\tau$ is a temperature parameter.

\paragraph{Prototype-to-Class Loss.}
The first objective aligns each prototype with the hub of its own class:
\begin{equation}
    \mathcal L_{\mathrm{P2C}}
    =
    -\frac{1}{2R}
    \sum_{c=0}^{1}
    \sum_{r=1}^{R}
    \log
    \frac{
        \phi(p_{c,r},h_c)
    }{
        \sum_{d=0}^{1}\phi(p_{c,r},h_d)
    }.
    \label{eq:p2c}
\end{equation}
This objective, whose denominator sums over the two class hubs, preserves the class-level orientation of each prototype bank.

\paragraph{Sample-to-Prototype Loss.}
We compute a soft assignment over the prototypes of the ground-truth class:
\begin{equation}
    q_{i,r}
    =
    \frac{
        \phi(z_i,p_{y_i,r})
    }{
        \sum_{k=1}^{R}\phi(z_i,p_{y_i,k})
    }.
    \label{eq:assignment}
\end{equation}
The assignment is treated as a stop-gradient target in the Sample-to-Prototype loss:
\begin{equation}
\begin{aligned}
    \mathcal L_{\mathrm{S2P}}
    &=
    -\frac{1}{|B|}
    \sum_{i\in B}
    \sum_{r=1}^{R}
    \mathrm{sg}(q_{i,r})
    \\
    &\quad\cdot
    \log
    \frac{
        \phi(z_i,p_{y_i,r})
    }{
        \sum_{p\in\mathcal P}\phi(z_i,p)
    }.
\end{aligned}
    \label{eq:s2p}
\end{equation}
Here, $\mathrm{sg}(\cdot)$ denotes the stop-gradient operator.
This objective associates each sample with its ground-truth prototype bank while separating it from the opposite-class bank.

\paragraph{Prototype Positioning Loss.}
Sample-to-Prototype associates samples with prototypes but does not provide a prototype-specific target for within-class variation.
We therefore remove the class-hub component from the samples and prototypes:
\begin{equation}
    z_i^\perp
    =
    z_i
    -
    (z_i^\top h_{y_i})h_{y_i},
    \label{eq:sample-residual}
\end{equation}
\begin{equation}
    p_{c,r}^\perp
    =
    \mathrm{norm}
    \left(
        p_{c,r}
        -
        (p_{c,r}^\top h_c)h_c
    \right).
    \label{eq:prototype-residual}
\end{equation}
Here, $p_{c,r}^{\perp}$ is the normalized residual vector of prototype $p_{c,r}$ after removing its class-hub component.

Using the assignments in Equation~\ref{eq:assignment}, we construct a residual aggregate for each prototype:
\begin{equation}
    g_{c,r}^{\perp}
    =
    \sum_{i\in B_c}
    q_{i,r}z_i^\perp,
    \qquad
    \bar z_{c,r}^{\perp}
    =
    \mathrm{norm}
    \left(g_{c,r}^{\perp}\right).
    \label{eq:residual-aggregate}
\end{equation}
The vector $\bar z_{c,r}^{\perp}$ summarizes the residual variation of class-$c$ samples associated with prototype $p_{c,r}$.

Let
$\mathcal P^\perp=\{p_{d,k}^\perp\}_{d,k}$
denote the set of prototype residual vectors.
The Prototype Positioning loss is
\begin{equation}
    \mathcal L_{\mathrm{PP}}
    =
    -\frac{1}{2R}
    \sum_{c=0}^{1}
    \sum_{r=1}^{R}
    \log
    \frac{
        \phi(\bar z_{c,r}^{\perp},p_{c,r}^{\perp})
    }{
        \sum_{p^\perp\in\mathcal P^\perp}
        \phi(\bar z_{c,r}^{\perp},p^\perp)
    }.
    \label{eq:pp}
\end{equation}
Equation~\ref{eq:pp} is a softmax cross-entropy over the prototype residual vectors, with a separate data-derived target constructed for each prototype.
Residual vectors from both classes appear in the denominator and therefore compete in the shared embedding space.
Because $p_{c,r}^{\perp}$ is normalized, Prototype Positioning controls the direction of the residual component.
Prototype-to-Class preserves class alignment, and Sample-to-Prototype connects the resulting prototype to same-class samples.

\paragraph{Final Training Objective.}
The final objective is
\begin{equation}
    \mathcal L_{\mathrm{train}}
    =
    \mathcal L_{\mathrm{P2C}}
    +
    \mathcal L_{\mathrm{S2P}}
    +
    \mathcal L_{\mathrm{PP}}.
    \label{eq:training-objective}
\end{equation}
We jointly optimize the encoder and prototype parameters and renormalize the prototypes after each update.

\subsection{Inference}

Given an input text $x$, we compute
$z=\mathrm{norm}(f_\theta(x))$
and score each class by its most similar prototype:
\begin{equation}
    s_c(z)
    =
    \max_{r\in\{1,\ldots,R\}}
    z^\top p_{c,r}.
    \label{eq:class-score}
\end{equation}
The detection score and prediction are
\begin{equation}
    S(z)
    =
    s_1(z)-s_0(z),
    \qquad
    \hat y
    =
    \mathbf{1}\{S(z)>\delta\}.
    \label{eq:detection-score}
\end{equation}
Appendix~\ref{app:weighted_inference} evaluates a weighted within-class alternative using the same frozen model parameters.
\section{Experiments}
\label{sec:experiments}

\begin{table*}[t]
\centering
{\normalsize
\renewcommand{\arraystretch}{1.18}
\setlength{\tabcolsep}{5.5pt}
\begin{tabular*}{\textwidth}{@{\extracolsep{\fill}}lccc@{}}
\hline
Method & MAGE CDCM & RAID & M4 \\
\hline
Binary CE
&
\shortstack[c]{90.79\\{\footnotesize (83.18/98.40)}}
&
\shortstack[c]{86.81\\{\footnotesize (74.60/99.02)}}
&
\shortstack[c]{76.68\\{\footnotesize (54.56/98.80)}}
\\
SupCon
&
\shortstack[c]{94.77\\{\footnotesize (92.98/96.56)}}
&
\shortstack[c]{77.67\\{\footnotesize (58.94/96.39)}}
&
\shortstack[c]{83.84\\{\footnotesize (72.37/95.30)}}
\\
DeTeCtive
&
\shortstack[c]{\underline{94.84}\\{\footnotesize (91.87/97.81)}}
&
\shortstack[c]{\underline{87.68}\\{\footnotesize (80.17/95.20)}}
&
\shortstack[c]{\textbf{92.74}\\{\footnotesize (87.92/97.57)}}
\\
DSVDD
&
\shortstack[c]{94.43\\{\footnotesize (95.18/93.67)}}
&
\shortstack[c]{86.17\\{\footnotesize (76.36/95.97)}}
&
\shortstack[c]{81.08\\{\footnotesize (63.13/99.03)}}
\\
\hline
\textbf{MD-ProTector}
&
\shortstack[c]{\textbf{95.14}\\{\footnotesize (95.81/94.47)}}
&
\shortstack[c]{\textbf{88.18}\\{\footnotesize (82.52/93.84)}}
&
\shortstack[c]{\underline{86.03}\\{\footnotesize (76.87/95.20)}}
\\
\hline
\end{tabular*}
}
\caption{
Benchmark-level evaluation results.
Each cell reports AvgRec with HumanRec/MachineRec in parentheses.
MAGE CDCM evaluates mixed domain and generator conditions, RAID evaluates adversarial and decoding robustness, and M4 evaluates language shift.
Bold and underline indicate the best and second-best AvgRec within each setting, respectively.
}
\label{tab:benchmark_results}
\end{table*}

\begin{table*}[t]
\centering
{\normalsize
\renewcommand{\arraystretch}{1.18}
\setlength{\tabcolsep}{7.0pt}
\begin{tabular*}{\textwidth}{@{\extracolsep{\fill}}lcc@{}}
\hline
Method & MAGE Unseen Domains & MAGE Unseen Models \\
\hline
Binary CE
&
\shortstack[c]{67.99\\{\footnotesize (37.04/98.94)}}
&
\shortstack[c]{89.78\\{\footnotesize (85.85/93.71)}}
\\
SupCon
&
\shortstack[c]{75.11\\{\footnotesize (56.82/93.41)}}
&
\shortstack[c]{90.92\\{\footnotesize (93.49/88.34)}}
\\
DeTeCtive
&
\shortstack[c]{76.72\\{\footnotesize (55.75/97.69)}}
&
\shortstack[c]{\textbf{91.69}\\{\footnotesize (92.08/91.30)}}
\\
DSVDD
&
\shortstack[c]{\textbf{79.08}\\{\footnotesize (63.46/94.71)}}
&
\shortstack[c]{90.71\\{\footnotesize (95.22/86.19)}}
\\
\hline
\textbf{MD-ProTector}
&
\shortstack[c]{\underline{78.59}\\{\footnotesize (61.46/95.72)}}
&
\shortstack[c]{\underline{91.34}\\{\footnotesize (95.63/87.05)}}
\\
\hline
\end{tabular*}
}
\caption{
MAGE leave-one-out evaluation results.
Each cell reports AvgRec with HumanRec/MachineRec in parentheses.
MAGE Unseen Domains and MAGE Unseen Models report averages over 10 independently trained leave-one-out-domain scenarios and 7 independently trained leave-one-generator-family-out scenarios, respectively.
Bold and underline indicate the best and second-best AvgRec within each setting.
}
\label{tab:mage_leaveoneout_results}
\end{table*}

\subsection{Experimental Setup}

\paragraph{Evaluation Settings.}
We evaluate MD-ProTector on MAGE, RAID, and M4, which cover distinct deployment scenarios. 
MAGE is used to evaluate domain and generator generalization: MAGE Cross-Domain Cross-Model (MAGE CDCM) tests mixed domain/generator conditions, while MAGE Unseen Domains and MAGE Unseen Models test leave-one-out domain and leave-one-out generator-family generalization \citep{li-etal-2024-mage}. 
M4 evaluates multilingual and unseen-language generalization \citep{wang2024m4}. 
RAID evaluates robustness to adversarial attacks and decoding-related variations \citep{dugan2024raid}. 
Each MAGE leave-one-out scenario trains a separate detector, yielding 10 domain-shift and 7 generator-family-shift evaluations.
Detailed dataset statistics and split construction are provided in Appendix~\ref{app:evaluation_protocol}.

\paragraph{Baselines.}
We compare input-only encoder detectors under a common data and model-access protocol.
Binary CE and SupCon provide standard classification and supervised contrastive references, while DeTeCtive and DSVDD introduce structured representation objectives through hierarchical contrastive learning with KNN inference and machine-class compactness, respectively \citep{guo2024detective,zeng2025human-ood}.
All methods use the same data splits, encoder backbone, training budget, checkpoint selection, and validation-based threshold selection.
The comparison isolates the detector formulation without access to model internals or auxiliary generation.

\paragraph{Evaluation Metrics.}
Following MAGE \citep{li-etal-2024-mage}, we use Average Recall (AvgRec), the mean of HumanRec and MachineRec, as the primary metric and report the values in the main tables.
AvgRec evaluates the two class recalls with equal weight at the fixed decision threshold, preventing high recall on one class from masking failure on the other.
We choose $\delta$ on the validation split to maximize AvgRec and keep it fixed on the test split.
Section~\ref{sec:evaluation-results} discusses AUROC and FPR95, while complete F1, Accuracy, AUROC, AUPR, FPR95, and per-scenario results are provided in Appendix~\ref{app:full_results}.

\paragraph{Implementation Details.}
Unless otherwise specified, all encoder-based methods use the same backbone encoder within each evaluation setting.
Pretrained encoder checkpoints are loaded from the HuggingFace model hub and used with the HuggingFace Transformers implementation \citep{wolf-etal-2020-transformers}.
We use 125M Unsupervised SimCSE-RoBERTa as the default encoder \citep{gao-etal-2021-simcse}. 
All models are trained with a batch size of 256 using AdamW with a learning rate of $2\times10^{-5}$. 
Models are trained for 30 epochs with 2,000 warmup steps. 
The checkpoint with the best AvgRec on the validation set is selected for evaluation.
For MD-ProTector, the number of prototypes per class is set to $R=8$. 
The main experiments are conducted on a single NVIDIA B200 GPU with mixed BF16 precision.

\begin{table}[!t]
\centering
\normalsize
\renewcommand{\arraystretch}{1.04}
\setlength{\tabcolsep}{0pt}
\begin{tabular*}{\columnwidth}{@{}l@{\extracolsep{\fill}}r@{}}
\hline
Variant & AvgRec \\
\hline
\multicolumn{2}{@{}l@{}}{\textit{Prototype Positioning} $(R=8,\ \tau=0.15)$} \\
Full objective & \textbf{95.14} \\
w/o $\mathcal{L}_{\mathrm{PP}}$ & 94.78 \\
PP $\rightarrow$ Simple Prototype Repulsion & 94.55 \\
Raw-space PP & 94.33 \\
\hline
\multicolumn{2}{@{}l@{}}{\textit{Prototype Initialization}} \\
K-Means & \textbf{95.14} \\
Random & 94.50 \\
\hline
\multicolumn{2}{@{}l@{}}{\textit{Number of Prototypes} $(\tau=0.15)$} \\
$R=1$ & 94.50 \\
$R=2$ & 95.03 \\
$R=4$ & 94.88 \\
$R=8$ & \textbf{95.14} \\
$R=16$ & 94.53 \\
$R=32$ & 94.36 \\
\hline
\multicolumn{2}{@{}l@{}}{\textit{Temperature} $(R=8)$} \\
$\tau=0.07$ & 94.91 \\
$\tau=0.10$ & 95.07 \\
$\tau=0.15$ & \textbf{95.14} \\
$\tau=0.20$ & 95.05 \\
$\tau=0.50$ & 93.97 \\
\hline
\multicolumn{2}{@{}l@{}}{\textit{Encoder Backbone}} \\
unsup-simcse-roberta-base & \textbf{95.14} \\
roberta-base & 94.66 \\
sup-simcse-roberta-base & 94.01 \\
e5-base & 93.70 \\
bert-base-uncased & 93.60 \\
unsup-simcse-bert-base & 93.24 \\
bge-base-en-v1.5 & 92.86 \\
\hline
\end{tabular*}
\caption{
Ablation studies on the MAGE CDCM dataset.
All entries report AvgRec.
The default configuration uses the full objective, $R=8$, $\tau=0.15$, K-Means initialization, and the unsupervised SimCSE-RoBERTa encoder.
``Raw-space PP'' denotes Prototype Positioning without removing the class hub direction.
}
\label{tab:ablation_results}
\end{table}

\subsection{Evaluation Results}
\label{sec:evaluation-results}

Tables~\ref{tab:benchmark_results} and~\ref{tab:mage_leaveoneout_results} report AvgRec together with the recall for each class under the five evaluation settings.

\paragraph{Mixed and Adversarial Conditions.}
MD-ProTector obtains the highest AvgRec on both MAGE CDCM and RAID.
On MAGE CDCM, it reaches 95.14 AvgRec with balanced recalls of 95.81 for human and 94.47 for machine text.
Its AUROC of 98.41 and FPR95 of 4.89 are also second-best, indicating that the AvgRec gain is accompanied by strong score-level separation under mixed domain and generator conditions.
On RAID, MD-ProTector achieves the highest AvgRec (88.18), HumanRec (82.52), and AUROC (95.41), together with the lowest FPR95 (27.78).

\paragraph{Held-Out Generator and Domain Shifts.}
Under unseen-generator-family evaluation, MD-ProTector obtains the second-highest AvgRec of 91.34 and the lowest FPR95 of 11.44. 
Its HumanRec of 95.63 is the highest among the evaluated methods, although MachineRec remains lower than that of DeTeCtive. 
Under unseen-domain evaluation, MD-ProTector again ranks second in AvgRec at 78.59, narrowly below DSVDD at 79.08. 
DSVDD retains stronger AUROC and FPR95 in this setting, showing that the prototype organization improves class-balanced performance more consistently than score ordering under every type of domain shift.

\paragraph{Language Shift.}
M4 remains the most challenging setting for MD-ProTector. 
It obtains the second-highest AvgRec of 86.03, improving HumanRec to 76.87 compared with 54.56 for Binary CE, 72.37 for SupCon, and 63.13 for DSVDD, while retaining 95.20 MachineRec. 
However, its AUROC and FPR95 remain below Binary CE and DSVDD. 
The remaining error is therefore concentrated in representing human-written text in unseen languages, rather than in detecting machine-generated text. 
Appendix~\ref{app:weighted_inference} reports that a weighted average of prototype similarities improves the frozen checkpoint AvgRec without retraining.
Across the five settings, MD-ProTector ranks within the top two in AvgRec.

\subsection{Ablation Studies}
\label{sec:deepfake_ablation}

Table~\ref{tab:ablation_results} examines Prototype Positioning together with the main configuration choices on MAGE CDCM.
The objective ablation retains Prototype-to-Class and Sample-to-Prototype and uses the same initialization, prototype count, and temperature across all variants.
Removing $\mathcal{L}_{\mathrm{PP}}$ lowers AvgRec from 95.14 to 94.78, and PP leads across three other seeds (Appendix~\ref{app:seed_variation}).
Replacing Prototype Positioning with simple prototype repulsion yields 94.55, while positioning prototypes without removing the class-hub direction yields 94.33.
These comparisons support the proposed formulation, in which each prototype is positioned using the residual variation of its associated samples.

K-Means initialization improves AvgRec from 94.50 to 95.14 relative to random initialization, indicating that the observed class distributions provide a useful starting point for optimization.
Increasing the number of prototypes from $R=1$ to $R=2$ raises AvgRec from 94.50 to 95.03, and reaches its maximum at $R=8$ and declines with larger prototype banks.
Once multiple prototypes are available, their organization remains important rather than capacity itself.

AvgRec remains between 94.91 and 95.14 for $\tau\in[0.07,0.20]$ and decreases to 93.97 at $\tau=0.50$.
Across encoder backbones, AvgRec ranges from 92.86 to 95.14, with unsupervised SimCSE-RoBERTa obtaining the highest value.

\subsection{Prototype Analysis}
\label{sec:prototype-analysis}

\begin{figure*}[t]
\centering
\begin{minipage}{0.49\linewidth}
    \centering
    \includegraphics[width=\linewidth]{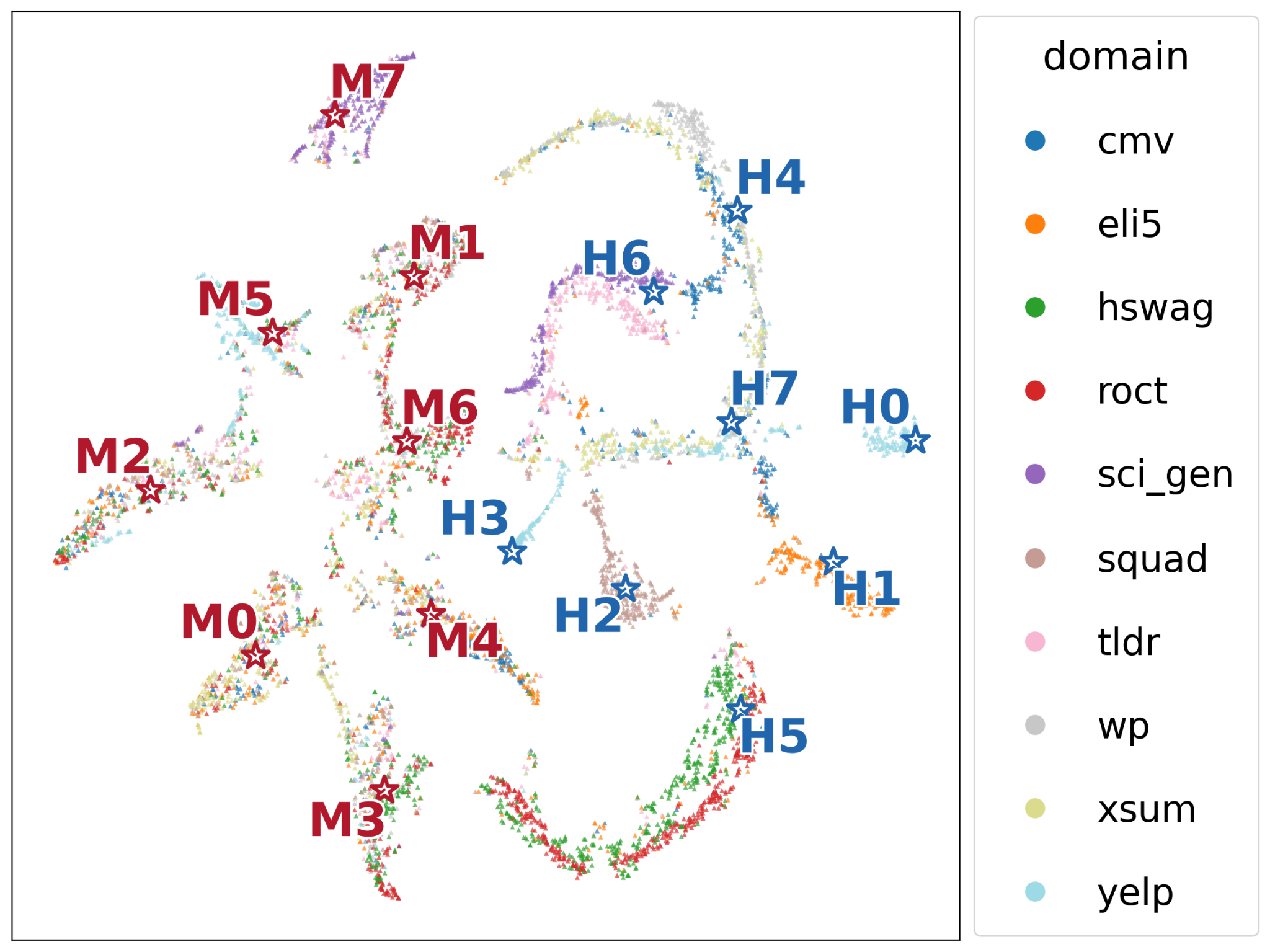}
    \centerline{\small (a) Colored by domain}
\end{minipage}
\hfill
\begin{minipage}{0.49\linewidth}
    \centering
    \includegraphics[width=\linewidth]{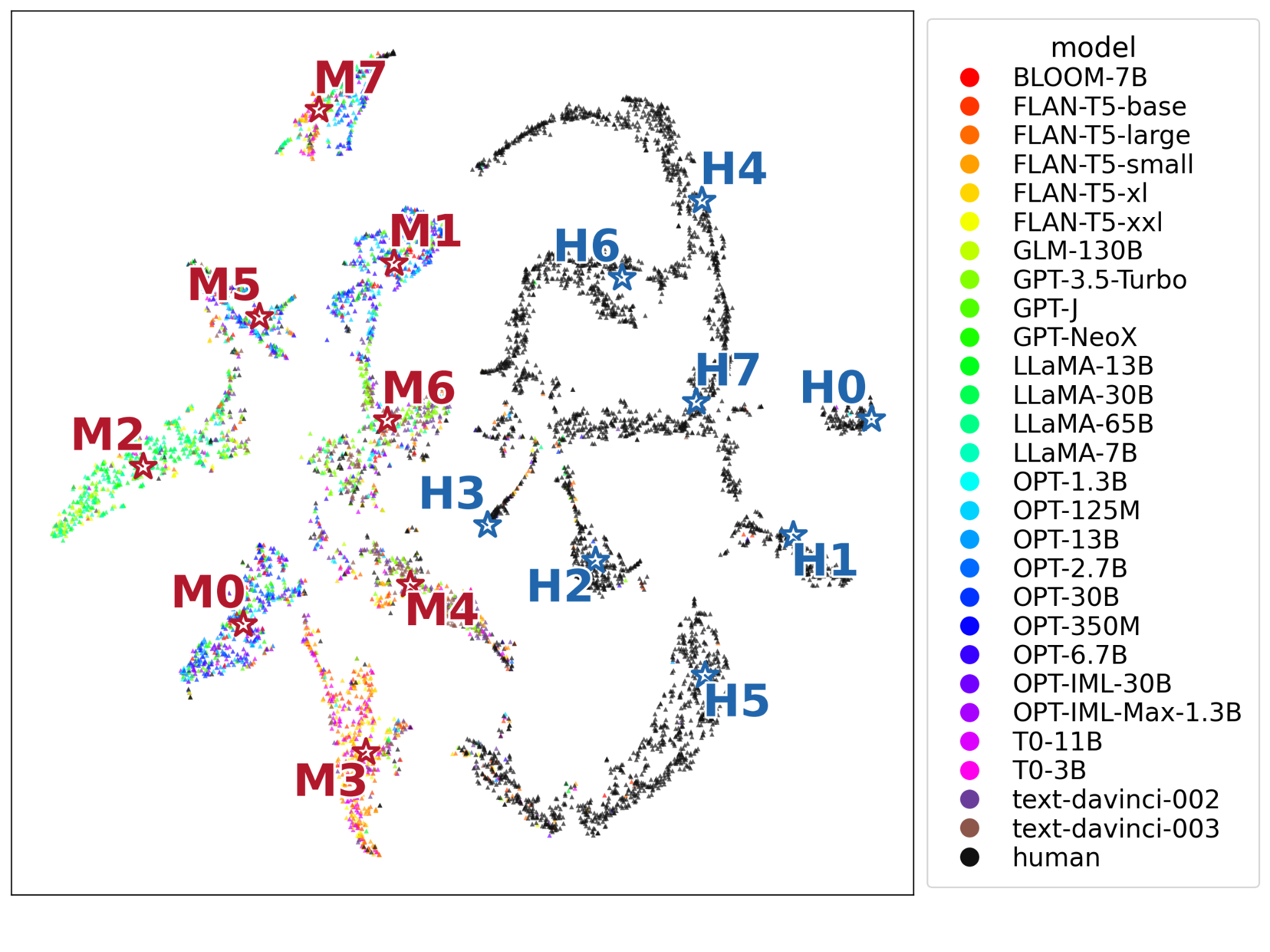}
    \centerline{\small (b) Colored by generator model}
\end{minipage}
\caption{
Prototype visualization on the MAGE CDCM dataset.
Both panels show the same t-SNE projection of normalized test embeddings and learned prototypes.
The left panel colors samples by domain, and the right panel colors samples by generator model.
Dots denote test samples.
Blue and red stars indicate human and machine prototypes, respectively.
}
\label{fig:tsne_deepfake}
\end{figure*}

\begin{table*}[t!]
\centering
\normalsize
\renewcommand{\arraystretch}{1.05}
\setlength{\tabcolsep}{4.3pt}

\textbf{(a) Machine prototypes}\par\vspace{0.15em}
\begin{tabular*}{\textwidth}{@{\extracolsep{\fill}}llll@{}}
\hline
Prototype ($n$) & Top domain & Top generator & Writing cues \\
\hline
\textbf{M1} (424) & SQuAD 19.8\% & OPT-6.7B 11.6\% & Quotation 77.6\%, Instruction 78.1\% \\
\textbf{M5} (386) & Yelp 62.7\% & BLOOM-7B 7.5\% & Review 58.8\% \\
\textbf{M7} (311) & SciGen 74.6\% & GPT-3.5-Turbo 6.4\% & Academic 49.2\% \\
\hline
\end{tabular*}

\vspace{0.45em}
\textbf{(b) Human prototypes}\par\vspace{0.15em}
\begin{tabular*}{\textwidth}{@{\extracolsep{\fill}}lll@{}}
\hline
Prototype ($n$) & Top domain & Writing cues \\
\hline
\textbf{H0} (147) & Yelp 93.2\% & First person 98.6\%, Review 85.7\% \\
\textbf{H6} (510) & TLDR 41.4\%, SciGen 40.6\% & Academic 33.5\% \\
\hline
\end{tabular*}

\caption{
Selected groups of texts assigned to prototypes.
Top domain and generator are the largest shares among assigned texts.
Writing-cue percentages are the fractions containing each cue.
Complete summaries and cue definitions are provided in
Appendix~\ref{app:prototype-analysis-details}.
}
\label{tab:main-prototype-groups}
\end{table*}

Figure~\ref{fig:tsne_deepfake} shows the learned prototypes distributed across multiple occupied regions of the human and machine embedding spaces.
All prototypes show their own test sample coverage, showing that both banks retain multiple active representatives and avoid complete assignment collapse.

We use M0--M7 and H0--H7 to identify the machine and human prototypes.
Table~\ref{tab:main-prototype-groups} summarizes selected groups characterized by instructional, review, and academic writing cues.
The same cues appear in both classes and describe recurring patterns within each class.
The largest generator share among the machine prototypes is 24.8\%, indicating that each group includes texts from multiple generators.
These results show that the prototype banks organize distinct same-class text groups across domain and generator boundaries.
With PP, prototype assignments cut held-out cue-prediction error by 20.16\%, versus 0.58\% without PP (Appendix~\ref{app:cue_prediction}).

\section{Conclusion}
\label{sec:conclusion}
  In this work, we introduced MD-ProTector, an input-only encoder detector that represents human-written and LLM-generated text with separate banks of trainable prototypes.
By separating class-level alignment from prototype-specific residual positioning, MD-ProTector organizes multiple prototypes using the variation observed within each class while retaining direct prototype-based inference.
Across five controlled settings, the method achieves the highest AvgRec on MAGE CDCM and RAID.
On RAID, it also attains the highest AUROC and lowest FPR95 among the compared methods.
Ablations on MAGE CDCM further show that residual positioning yields stronger performance than the single-prototype, direct-repulsion, and raw-space positioning variants.
These results support data-derived positioning as an effective mechanism for organizing multiple class prototypes for LLM-generated text detection.
\section*{Limitations}
\label{sec:limitations}

This work assumes a fixed-label binary detection setting in which each input is classified as either human-written or LLM-generated, and does not address more complex scenarios such as partial generation, human–machine co-editing, or estimating degrees of machine involvement.
Detecting human-written text from unseen domains also remains difficult for every compared detector, including ours, particularly on held-out ROCStories and XSum (Appendix~\ref{app:full_results}).
In addition, the number of prototypes per class is treated as an empirically fixed design choice.
Adaptive mechanisms for adjusting prototype cardinality based on data characteristics are not explored.
Finally, our experiments are conducted under fixed training, validation, and test splits. 
In practical deployment, the distribution of generators, prompts, writing styles, and adversarial perturbations may change over time. 
While MD-ProTector initializes and optimizes prototypes from training data, continual prototype adaptation under temporal distribution drift remains future work.
\section*{Ethics Statement}
\label{sec:ethics}

The datasets utilized do not include private data or non-public personally identifiable information. The development of reliable text detection systems is crucial for maintaining trust in digital information. However, we acknowledge that detection technologies can potentially be used in a dual-use manner—adversaries might use our detector as a discriminator to train more sophisticated generators that evade detection. Additionally, while we strove to use diverse datasets, the "Human" class in our training data is sourced from web texts (e.g., Reddit, Wikipedia), which may contain inherent biases. Users should be cautious when deploying this model in sensitive contexts, as false positives could unfairly penalize human writers.
This risk is not uniform across inputs.
In our experiments, MD-ProTector's errors concentrate on human-written text from held-out domains and languages, with HumanRec of 13.92 on held-out ROCStories, 25.68 on held-out XSum, and 76.87 on M4 (Appendix~\ref{app:full_results}).
Detectors should therefore be validated on human-written text from the target domain or language before deployment.

\bibliography{custom}

@inproceedings{agrahari2025protofew,
  author = {Agrahari, Shifali and Kumar, Sujit and Sanasam, Ranbir Singh},
  title = {Can You Really Trust That Review? {P}roto{F}ew{R}o{BERT}a and {D}etect{AIR}ev: A Prototypical Few-Shot Method and Multi-Domain Benchmark for Detecting {AI}-Generated Reviews},
  booktitle = {Proceedings of the 14th International Joint Conference on Natural Language Processing and the 4th Conference of the Asia-Pacific Chapter of the Association for Computational Linguistics},
  pages = {2118--2140},
  month = dec,
  year = {2025},
  address = {Mumbai, India},
  publisher = {The Asian Federation of Natural Language Processing and The Association for Computational Linguistics},
  doi = {10.18653/v1/2025.findings-ijcnlp.132},
  url = {https://aclanthology.org/2025.findings-ijcnlp.132/},
}

@article{bakhtin-etal-2019-real,
  author = {Bakhtin, Anton and Gross, Sam and Ott, Myle and Deng, Yuntian and Ranzato, Marc'Aurelio and Szlam, Arthur},
  title = {Real or Fake? Learning to Discriminate Machine from Human Generated Text},
  journal = {arXiv preprint arXiv:1906.03351},
  year = {2019},
  url = {https://arxiv.org/abs/1906.03351},
}

@inproceedings{bao2024fastdetectgpt,
  author = {Bao, Guangsheng and Zhao, Yanbin and Teng, Zhiyang and Yang, Linyi and Zhang, Yue},
  title = {Fast-Detect{GPT}: Efficient Zero-Shot Detection of Machine-Generated Text via Conditional Probability Curvature},
  booktitle = {International Conference on Learning Representations},
  year = {2024},
  url = {https://openreview.net/forum?id=Bpcgcr8E8Z},
}

@article{chen2024proto,
  author = {Chen, Jun{-}Kun and Mei, Jilin and Chen, Liang and Zhao, Fangzhou and Xing, Yan and Hu, Yu},
  title = {{Proto-OOD}: Enhancing {OOD} Object Detection with Prototype Feature Similarity},
  journal = {arXiv preprint arXiv:2409.05466},
  year = {2024},
  url = {https://doi.org/10.48550/arXiv.2409.05466},
}

@article{chen2025imbd,
  author = {Chen, Jiaqi and Zhu, Xiaoye and Liu, Tianyang and Chen, Ying and Chen, Xinhui and Yuan, Yiwen and Leong, Chak Tou and Li, Zuchao and Tang, Long and Zhang, Lei and Yan, Chenyu and Mei, Guanghao and Zhang, Jie and Zhang, Lefei},
  title = {Imitate Before Detect: Aligning Machine Stylistic Preference for Machine-Revised Text Detection},
  journal = {Proceedings of the AAAI Conference on Artificial Intelligence},
  volume = {39},
  number = {22},
  pages = {23559--23567},
  year = {2025},
  doi = {10.1609/aaai.v39i22.34525},
}

@inproceedings{christ2024undetectable,
  author = {Christ, Miranda and Gunn, Sam and Zamir, Or},
  title = {Undetectable Watermarks for Language Models},
  booktitle = {Proceedings of Thirty Seventh Conference on Learning Theory},
  volume = {247},
  series = {Proceedings of Machine Learning Research},
  pages = {1125--1139},
  year = {2024},
  publisher = {PMLR},
  url = {https://proceedings.mlr.press/v247/christ24a.html},
}

@inproceedings{cui2016fine,
  author = {Cui, Yin and Zhou, Feng and Lin, Yuanqing and Belongie, Serge},
  title = {Fine-Grained Categorization and Dataset Bootstrapping Using Deep Metric Learning with Humans in the Loop},
  booktitle = {IEEE Conference on Computer Vision and Pattern Recognition},
  pages = {1153--1162},
  year = {2016},
  address = {Las Vegas, NV, USA},
  publisher = {{IEEE}},
  doi = {10.1109/CVPR.2016.130},
  url = {https://doi.org/10.1109/CVPR.2016.130},
}

@inproceedings{devlin-etal-2019-bert,
  author = {Devlin, Jacob and Chang, Ming-Wei and Lee, Kenton and Toutanova, Kristina},
  title = {{BERT}: Pre-training of Deep Bidirectional Transformers for Language Understanding},
  booktitle = {Proceedings of the 2019 Conference of the North American Chapter of the Association for Computational Linguistics: Human Language Technologies, Volume 1 (Long and Short Papers)},
  pages = {4171--4186},
  month = jun,
  year = {2019},
  address = {Minneapolis, Minnesota},
  publisher = {Association for Computational Linguistics},
  doi = {10.18653/v1/N19-1423},
  url = {https://aclanthology.org/N19-1423/},
}

@article{dong2024multi,
  author = {Dong, Zhijin and Liu, Hongzhi and Ren, Boyuan and Xiong, Weimin and Wu, Zhonghai},
  title = {Reconstruction-based Multi-Normal Prototypes Learning for Weakly Supervised Anomaly Detection},
  journal = {arXiv preprint arXiv:2408.14498},
  year = {2024},
  url = {https://arxiv.org/abs/2408.14498},
}

@inproceedings{dugan2024raid,
  author = {Dugan, Liam and Hwang, Alyssa and Trhl\'{\i}k, Filip and Zhu, Andrew and Ludan, Josh Magnus and Xu, Hainiu and Ippolito, Daphne and Callison-Burch, Chris},
  title = {{RAID}: A Shared Benchmark for Robust Evaluation of Machine-Generated Text Detectors},
  booktitle = {Proceedings of the 62nd Annual Meeting of the Association for Computational Linguistics (Volume 1: Long Papers)},
  pages = {12463--12492},
  month = aug,
  year = {2024},
  address = {Bangkok, Thailand},
  publisher = {Association for Computational Linguistics},
  doi = {10.18653/v1/2024.acl-long.674},
  url = {https://aclanthology.org/2024.acl-long.674/},
}

@inproceedings{fu2025detectanyllm,
  author = {Fu, Jiachen and Guo, Chun-Le and Li, Chongyi},
  title = {{DetectAnyLLM}: Towards Generalizable and Robust Detection of Machine-Generated Text Across Domains and Models},
  booktitle = {Proceedings of the 33rd ACM International Conference on Multimedia},
  pages = {11229--11238},
  year = {2025},
  doi = {10.1145/3746027.3754862},
}

@inproceedings{gao-etal-2021-simcse,
  author = {Gao, Tianyu and Yao, Xingcheng and Chen, Danqi},
  title = {{S}im{CSE}: Simple Contrastive Learning of Sentence Embeddings},
  booktitle = {Proceedings of the 2021 Conference on Empirical Methods in Natural Language Processing},
  pages = {6894--6910},
  month = nov,
  year = {2021},
  address = {Online and Punta Cana, Dominican Republic},
  publisher = {Association for Computational Linguistics},
  doi = {10.18653/v1/2021.emnlp-main.552},
  url = {https://aclanthology.org/2021.emnlp-main.552/},
}

@inproceedings{gehrmann-etal-2019-gltr,
  author = {Gehrmann, Sebastian and Strobelt, Hendrik and Rush, Alexander},
  title = {{GLTR}: Statistical Detection and Visualization of Generated Text},
  booktitle = {Proceedings of the 57th Annual Meeting of the Association for Computational Linguistics: System Demonstrations},
  pages = {111--116},
  month = jul,
  year = {2019},
  address = {Florence, Italy},
  publisher = {Association for Computational Linguistics},
  doi = {10.18653/v1/P19-3019},
  url = {https://aclanthology.org/P19-3019/},
}

@inproceedings{guo2024biscope,
  author = {Guo, Hanxi and Cheng, Siyuan and Jin, Xiaolong and Zhang, Zhuo and Zhang, Kaiyuan and Tao, Guanhong and Shen, Guangyu and Zhang, Xiangyu},
  title = {{BISCOPE}: {AI}-Generated Text Detection by Checking Memorization of Preceding Tokens},
  booktitle = {Advances in Neural Information Processing Systems},
  volume = {37},
  pages = {104065--104090},
  year = {2024},
  publisher = {Curran Associates, Inc.},
  doi = {10.52202/079017-3307},
  url = {https://proceedings.neurips.cc/paper_files/paper/2024/hash/bc808cf2d2444b0abcceca366b771389-Abstract-Conference.html},
}

@inproceedings{guo2024detective,
  author = {Guo, Xun and Zhang, Shan and He, Yongxin and Zhang, Ting and Feng, Wanquan and Huang, Haibin and Ma, Chongyang},
  title = {De{T}e{C}tive: Detecting {AI}-Generated Text via Multi-Level Contrastive Learning},
  booktitle = {Advances in Neural Information Processing Systems},
  volume = {37},
  pages = {88320--88347},
  year = {2024},
  publisher = {Curran Associates, Inc.},
  doi = {10.52202/079017-2802},
  url = {https://proceedings.neurips.cc/paper_files/paper/2024/hash/a117a3cd54b7affad04618c77c2fb18b-Abstract-Conference.html},
}

@inproceedings{hans2024binoculars,
  author = {Hans, Abhimanyu and Schwarzschild, Avi and Cherepanova, Valeriia and Kazemi, Hamid and Saha, Aniruddha and Goldblum, Micah and Geiping, Jonas and Goldstein, Tom},
  title = {Spotting {LLM}s With Binoculars: Zero-Shot Detection of Machine-Generated Text},
  booktitle = {Proceedings of the 41st International Conference on Machine Learning},
  volume = {235},
  series = {Proceedings of Machine Learning Research},
  pages = {17519--17537},
  year = {2024},
  publisher = {PMLR},
  url = {https://proceedings.mlr.press/v235/hans24a.html},
}

@inproceedings{hao2025learning2rewrite,
  author = {Hao, Wei and Li, Ran and Zhao, Weiliang and Yang, Junfeng and Mao, Chengzhi},
  title = {Learning to Rewrite: Generalized {LLM}-Generated Text Detection},
  booktitle = {Proceedings of the 63rd Annual Meeting of the Association for Computational Linguistics (Volume 1: Long Papers)},
  pages = {6421--6434},
  month = jul,
  year = {2025},
  address = {Vienna, Austria},
  publisher = {Association for Computational Linguistics},
  doi = {10.18653/v1/2025.acl-long.322},
  url = {https://aclanthology.org/2025.acl-long.322/},
}

@article{hong2023protorynet,
  author = {Hong, Dat and Wang, Tong and Baek, Stephen},
  title = {{ProtoryNet}: Interpretable Text Classification via Prototype Trajectories},
  journal = {Journal of Machine Learning Research},
  volume = {24},
  number = {264},
  pages = {1--39},
  year = {2023},
  url = {http://jmlr.org/papers/v24/21-0899.html},
}

@inproceedings{hu-etal-2023-radar,
  author = {Hu, Xiaomeng and Chen, Pin{-}Yu and Ho, Tsung{-}Yi},
  title = {{RADAR}: Robust {AI}-Text Detection via Adversarial Learning},
  booktitle = {Advances in Neural Information Processing Systems},
  volume = {36},
  year = {2023},
  publisher = {Curran Associates, Inc.},
  url = {https://proceedings.neurips.cc/paper_files/paper/2023/hash/30e15e5941ae0cdab7ef58cc8d59a4ca-Abstract-Conference.html},
}

@inproceedings{ippolito-etal-2020-automatic,
  author = {Ippolito, Daphne and Duckworth, Daniel and Callison-Burch, Chris and Eck, Douglas},
  title = {Automatic Detection of Generated Text is Easiest when Humans are Fooled},
  booktitle = {Proceedings of the 58th Annual Meeting of the Association for Computational Linguistics},
  pages = {1808--1822},
  month = jul,
  year = {2020},
  address = {Online},
  publisher = {Association for Computational Linguistics},
  doi = {10.18653/v1/2020.acl-main.164},
  url = {https://aclanthology.org/2020.acl-main.164/},
}

@article{jia2025diversified,
  author = {Jia, Yulong and Li, Jiaming and Zhao, Ganlong and Liu, Shuangyin and Sun, Weijun and Lin, Liang and Li, Guanbin},
  title = {Enhancing out-of-distribution detection via diversified multi-prototype contrastive learning},
  journal = {Pattern Recognition},
  volume = {161},
  pages = {111214},
  year = {2025},
  doi = {10.1016/j.patcog.2024.111214},
}

@inproceedings{kuznetsov-etal-2024-robust,
  author = {Kuznetsov, Kristian and Tulchinskii, Eduard and Kushnareva, Laida and Magai, German and Barannikov, Serguei and Nikolenko, Sergey and Piontkovskaya, Irina},
  title = {Robust {AI}-Generated Text Detection by Restricted Embeddings},
  booktitle = {Findings of the Association for Computational Linguistics: EMNLP 2024},
  pages = {17036--17055},
  month = nov,
  year = {2024},
  address = {Miami, Florida, USA},
  publisher = {Association for Computational Linguistics},
  doi = {10.18653/v1/2024.findings-emnlp.992},
  url = {https://aclanthology.org/2024.findings-emnlp.992/},
}

@article{kwon_2025_survey,
  author = {Kwon, Soonchan and Jang, Beakcheol},
  title = {A Comprehensive Survey of Fake Text Detection on Misinformation and {LM}-Generated Texts},
  journal = {IEEE Access},
  volume = {13},
  pages = {25301--25324},
  year = {2025},
  doi = {10.1109/ACCESS.2025.3538805},
}

@inproceedings{li-etal-2024-mage,
  author = {Li, Yafu and Li, Qintong and Cui, Leyang and Bi, Wei and Wang, Zhilin and Wang, Longyue and Yang, Linyi and Shi, Shuming and Zhang, Yue},
  title = {{MAGE}: Machine-generated Text Detection in the Wild},
  booktitle = {Proceedings of the 62nd Annual Meeting of the Association for Computational Linguistics (Volume 1: Long Papers)},
  pages = {36--53},
  month = aug,
  year = {2024},
  address = {Bangkok, Thailand},
  publisher = {Association for Computational Linguistics},
  doi = {10.18653/v1/2024.acl-long.3},
  url = {https://aclanthology.org/2024.acl-long.3/},
}

@inproceedings{liprototypical,
  author = {Li, Junnan and Zhou, Pan and Xiong, Caiming and Hoi, Steven},
  title = {Prototypical Contrastive Learning of Unsupervised Representations},
  booktitle = {International Conference on Learning Representations},
  year = {2021},
  url = {https://openreview.net/forum?id=KmykpuSrjcq},
}

@article{liu2019roberta,
  author = {Liu, Yinhan and Ott, Myle and Goyal, Naman and Du, Jingfei and Joshi, Mandar and Chen, Danqi and Levy, Omer and Lewis, Mike and Zettlemoyer, Luke and Stoyanov, Veselin},
  title = {{RoBERTa}: A Robustly Optimized {BERT} Pretraining Approach},
  journal = {arXiv preprint arXiv:1907.11692},
  year = {2019},
  url = {https://arxiv.org/abs/1907.11692},
}

@inproceedings{lu2024palm,
  author = {Lu, Haodong and Gong, Dong and Wang, Shuo and Xue, Jason and Yao, Lina and Moore, Kristen},
  title = {Learning with Mixture of Prototypes for Out-of-Distribution Detection},
  booktitle = {International Conference on Learning Representations},
  year = {2024},
  url = {https://openreview.net/forum?id=uNkKaD3MCs},
}

@inproceedings{mao2024raidar,
  author = {Mao, Chengzhi and Vondrick, Carl and Wang, Hao and Yang, Junfeng},
  title = {{RAIDAR}: Generative {AI} Detection via Rewriting},
  booktitle = {International Conference on Learning Representations},
  year = {2024},
  url = {https://openreview.net/forum?id=bQWE2UqXmf},
}

@inproceedings{mcqueen1967some,
  author = {MacQueen, J.},
  title = {Some Methods for Classification and Analysis of Multivariate Observations},
  booktitle = {Proceedings of the Fifth Berkeley Symposium on Mathematical Statistics and Probability},
  pages = {281--297},
  year = {1967},
  url = {https://digicoll.lib.berkeley.edu/record/113015/files/math_s5_v1_article-17},
}

@article{miralles2025pawn,
  author = {Miralles-Gonz\'{a}lez, Pablo and Huertas-Tato, Javier and Mart\'{\i}n, Alejandro and Camacho, David},
  title = {Not All Tokens Are Created Equal: Perplexity Attention Weighted Networks for {AI}-Generated Text Detection},
  journal = {Information Fusion},
  volume = {125},
  pages = {103465},
  year = {2026},
  doi = {10.1016/j.inffus.2025.103465},
}

@inproceedings{mitchell2023detectgpt,
  author = {Mitchell, Eric and Lee, Yoonho and Khazatsky, Alexander and Manning, Christopher D. and Finn, Chelsea},
  title = {Detect{GPT}: Zero-Shot Machine-Generated Text Detection using Probability Curvature},
  booktitle = {Proceedings of the 40th International Conference on Machine Learning},
  volume = {202},
  series = {Proceedings of Machine Learning Research},
  pages = {24950--24962},
  year = {2023},
  publisher = {PMLR},
  url = {https://proceedings.mlr.press/v202/mitchell23a.html},
}

@article{najjar2025detecting,
  author = {Najjar, Ayat A. and Ashqar, Huthaifa I. and Darwish, Omar A. and Hammad, Eman M.},
  title = {Detecting AI-Generated Text in Educational Content: Leveraging Machine Learning and Explainable {AI} for Academic Integrity},
  journal = {arXiv preprint arXiv:2501.03203},
  year = {2025},
  url = {https://arxiv.org/abs/2501.03203},
}

@inproceedings{pmlr-v202-kirchenbauer23a,
  author = {Kirchenbauer, John and Geiping, Jonas and Wen, Yuxin and Katz, Jonathan and Miers, Ian and Goldstein, Tom},
  title = {A Watermark for Large Language Models},
  booktitle = {Proceedings of the 40th International Conference on Machine Learning},
  volume = {202},
  series = {Proceedings of Machine Learning Research},
  pages = {17061--17084},
  month = jul,
  year = {2023},
  publisher = {PMLR},
  url = {https://proceedings.mlr.press/v202/kirchenbauer23a.html},
}

@inproceedings{rodriguez-etal-2022-cross,
  author = {Rodriguez, Juan Diego and Hay, Todd and Gros, David and Shamsi, Zain and Srinivasan, Ravi},
  title = {Cross-Domain Detection of {GPT}-2-Generated Technical Text},
  booktitle = {Proceedings of the 2022 Conference of the North American Chapter of the Association for Computational Linguistics: Human Language Technologies},
  pages = {1213--1233},
  month = jul,
  year = {2022},
  address = {Seattle, United States},
  publisher = {Association for Computational Linguistics},
  doi = {10.18653/v1/2022.naacl-main.88},
  url = {https://aclanthology.org/2022.naacl-main.88/},
}

@inproceedings{snell2017prototypical,
  author = {Snell, Jake and Swersky, Kevin and Zemel, Richard},
  title = {Prototypical Networks for Few-Shot Learning},
  booktitle = {Advances in Neural Information Processing Systems},
  volume = {30},
  pages = {4077--4087},
  year = {2017},
  publisher = {Curran Associates, Inc.},
  url = {https://proceedings.neurips.cc/paper_files/paper/2017/hash/cb8da6767461f2812ae4290eac7cbc42-Abstract.html},
}

@inproceedings{uchendu-etal-2020-authorship,
  author = {Uchendu, Adaku and Le, Thai and Shu, Kai and Lee, Dongwon},
  title = {Authorship Attribution for Neural Text Generation},
  booktitle = {Proceedings of the 2020 Conference on Empirical Methods in Natural Language Processing},
  pages = {8384--8395},
  month = nov,
  year = {2020},
  address = {Online},
  publisher = {Association for Computational Linguistics},
  doi = {10.18653/v1/2020.emnlp-main.673},
  url = {https://aclanthology.org/2020.emnlp-main.673/},
}

@inproceedings{verma2024ghostbuster,
  author = {Verma, Vivek and Fleisig, Eve and Tomlin, Nicholas and Klein, Dan},
  title = {Ghostbuster: Detecting Text Ghostwritten by Large Language Models},
  booktitle = {Proceedings of the 2024 Conference of the North American Chapter of the Association for Computational Linguistics: Human Language Technologies (Volume 1: Long Papers)},
  pages = {1702--1717},
  month = jun,
  year = {2024},
  address = {Mexico City, Mexico},
  publisher = {Association for Computational Linguistics},
  doi = {10.18653/v1/2024.naacl-long.95},
  url = {https://aclanthology.org/2024.naacl-long.95/},
}

@article{wang-etal-2023-implementing,
  author = {Wang, Zecong and Cheng, Jiaxi and Cui, Chen and Yu, Chenhao},
  title = {Implementing {BERT} and Fine-Tuned {RoBERTa} to Detect {AI}-Generated News by {ChatGPT}},
  journal = {arXiv preprint arXiv:2306.07401},
  year = {2023},
  url = {https://doi.org/10.48550/arXiv.2306.07401},
}

@article{wang2022text,
  author = {Wang, Liang and Yang, Nan and Huang, Xiaolong and Jiao, Binxing and Yang, Linjun and Jiang, Daxin and Majumder, Rangan and Wei, Furu},
  title = {Text Embeddings by Weakly-Supervised Contrastive Pre-training},
  journal = {arXiv preprint arXiv:2212.03533},
  year = {2022},
  url = {https://arxiv.org/abs/2212.03533},
}

@inproceedings{wang2024m4,
  author = {Wang, Yuxia and Mansurov, Jonibek and Ivanov, Petar and Su, Jinyan and Shelmanov, Artem and Tsvigun, Akim and Whitehouse, Chenxi and Mohammed Afzal, Osama and Mahmoud, Tarek and Sasaki, Toru and Arnold, Thomas and Aji, Alham Fikri and Habash, Nizar and Gurevych, Iryna and Nakov, Preslav},
  title = {{M4}: Multi-generator, Multi-domain, and Multilingual Black-Box Machine-Generated Text Detection},
  booktitle = {Proceedings of the 18th Conference of the European Chapter of the Association for Computational Linguistics (Volume 1: Long Papers)},
  pages = {1369--1407},
  month = mar,
  year = {2024},
  address = {St. Julian{'}s, Malta},
  publisher = {Association for Computational Linguistics},
  doi = {10.18653/v1/2024.eacl-long.83},
  url = {https://aclanthology.org/2024.eacl-long.83/},
}

@inproceedings{wolf-etal-2020-transformers,
  author = {Wolf, Thomas and Debut, Lysandre and Sanh, Victor and Chaumond, Julien and Delangue, Clement and Moi, Anthony and Cistac, Pierric and Rault, Tim and Louf, Remi and Funtowicz, Morgan and Davison, Joe and Shleifer, Sam and von Platen, Patrick and Ma, Clara and Jernite, Yacine and Plu, Julien and Xu, Canwen and Le Scao, Teven and Gugger, Sylvain and Drame, Mariama and Lhoest, Quentin and Rush, Alexander},
  title = {Transformers: State-of-the-Art Natural Language Processing},
  booktitle = {Proceedings of the 2020 Conference on Empirical Methods in Natural Language Processing: System Demonstrations},
  pages = {38--45},
  month = oct,
  year = {2020},
  address = {Online},
  publisher = {Association for Computational Linguistics},
  doi = {10.18653/v1/2020.emnlp-demos.6},
  url = {https://aclanthology.org/2020.emnlp-demos.6/},
}

@article{wu-etal-2025-survey,
  author = {Wu, Junchao and Yang, Shu and Zhan, Runzhe and Yuan, Yulin and Chao, Lidia Sam and Wong, Derek Fai},
  title = {A Survey on {LLM}-Generated Text Detection: Necessity, Methods, and Future Directions},
  journal = {Computational Linguistics},
  volume = {51},
  number = {1},
  pages = {275--338},
  month = mar,
  year = {2025},
  address = {Cambridge, MA},
  publisher = {MIT Press},
  doi = {10.1162/coli_a_00549},
  url = {https://aclanthology.org/2025.cl-1.8/},
}

@inproceedings{wu2024detectrl,
  author = {Wu, Junchao and Zhan, Runzhe and Wong, Derek and Yang, Shu and Yang, Xinyi and Yuan, Yulin and Chao, Lidia},
  title = {Detect{RL}: Benchmarking {LLM}-Generated Text Detection in Real-World Scenarios},
  booktitle = {Advances in Neural Information Processing Systems},
  volume = {37},
  pages = {100369--100401},
  year = {2024},
  publisher = {Curran Associates, Inc.},
  doi = {10.52202/079017-3186},
  url = {https://proceedings.neurips.cc/paper_files/paper/2024/hash/b61bdf7e9f64c04ec75a26e781e2ad51-Abstract-Datasets_and_Benchmarks_Track.html},
}

@inproceedings{wu2025moses,
  author = {Wu, Junxi and Wang, Jinpeng and Liu, Zheng and Chen, Bin and Hu, Dongjian and Wu, Hao and Xia, Shu-Tao},
  title = {{MoSEs}: Uncertainty-Aware {AI}-Generated Text Detection via Mixture of Stylistics Experts with Conditional Thresholds},
  booktitle = {Proceedings of the 2025 Conference on Empirical Methods in Natural Language Processing},
  pages = {5786--5805},
  month = nov,
  year = {2025},
  address = {Suzhou, China},
  publisher = {Association for Computational Linguistics},
  doi = {10.18653/v1/2025.emnlp-main.294},
  url = {https://aclanthology.org/2025.emnlp-main.294/},
}

@inproceedings{xiao2024cpack,
  author = {Xiao, Shitao and Liu, Zheng and Zhang, Peitian and Muennighoff, Niklas and Lian, Defu and Nie, Jian-Yun},
  title = {{C-Pack}: Packed Resources for General {Chinese} Embeddings},
  booktitle = {Proceedings of the 47th International ACM SIGIR Conference on Research and Development in Information Retrieval},
  pages = {641--649},
  year = {2024},
  address = {Washington, DC, USA},
  publisher = {Association for Computing Machinery},
  doi = {10.1145/3626772.3657878},
}

@article{xu2026samp,
  author = {Xu, Yan and Yang, Wenzhong and Yin, Yabo and Lv, Hongzhen and Wang, Zhenhua and He, Jingfeng and Jia, Xiangyi and Wang, Xianfeng},
  title = {{SAMP}: Source-Aware Multi-Prototype Learning for Machine-Generated Text Detection},
  journal = {Research Square},
  month = may,
  year = {2026},
  doi = {10.21203/rs.3.rs-9598516/v1},
  url = {https://doi.org/10.21203/rs.3.rs-9598516/v1},
  note = {Preprint},
}

@inproceedings{zeng2025human-ood,
  author = {Zeng, Cong and Tang, Shengkun and Chen, Yuanzhou and Shen, Zhiqiang and Yu, Wenchao and Zhao, Xujiang and Chen, Haifeng and Cheng, Wei and Xu, Zhiqiang},
  title = {Human Texts Are Outliers: Detecting {LLM}-Generated Texts via Out-of-Distribution Detection},
  booktitle = {Advances in Neural Information Processing Systems},
  volume = {38},
  year = {2025},
  publisher = {Curran Associates, Inc.},
  url = {https://proceedings.neurips.cc/paper_files/paper/2025/hash/ef52fd1e24634cb8f7003ebbfb3644d9-Abstract-Conference.html},
}

@article{zhang2025multiprototype,
  author = {Zhang, Yue and Chen, Yiyi and Fang, Chaowei and Wang, Qian and Wu, Jiayi and Xin, Jingmin},
  title = {Learning from open-set noisy labels based on multi-prototype modeling},
  journal = {Pattern Recognition},
  volume = {157},
  pages = {110902},
  year = {2025},
  doi = {10.1016/j.patcog.2024.110902},
}

\appendix
\section{Evaluation Set and Protocol Details}
\label{app:evaluation_protocol}

This section reports dataset statistics and protocol details for MAGE, M4, and RAID.
All datasets are cast as binary human-written versus machine-generated text detection, and human-written text is treated as the positive class in our metric implementation.
Validation data are used only to select checkpoints, thresholds, and hyperparameters. Test labels are not used for any of these choices.

\begin{table*}[t!]
\centering
\small
\renewcommand{\arraystretch}{1.08}
\setlength{\tabcolsep}{3pt}
\begin{minipage}[t]{0.31\textwidth}
\centering
\begin{tabular*}{\linewidth}{@{\extracolsep{\fill}}lrrr@{}}
\hline
\multicolumn{4}{@{}l}{\textit{MAGE CDCM}} \\
\hline
Split & Total & Human & Machine \\
\hline
Train & 319,071 & 93,318 & 225,753 \\
Valid & 56,792 & 28,799 & 27,993 \\
Test & 56,819 & 28,741 & 28,078 \\
\hline
\end{tabular*}
\end{minipage}
\hfill
\begin{minipage}[t]{0.31\textwidth}
\centering
\begin{tabular*}{\linewidth}{@{\extracolsep{\fill}}lrrr@{}}
\hline
\multicolumn{4}{@{}l}{\textit{M4 Multilingual}} \\
\hline
Split & Total & Human & Machine \\
\hline
Train & 172,417 & 83,846 & 88,571 \\
Valid & 4,000 & 2,000 & 2,000 \\
Test & 42,378 & 20,238 & 22,140 \\
\hline
\end{tabular*}
\end{minipage}
\hfill
\begin{minipage}[t]{0.31\textwidth}
\centering
\begin{tabular*}{\linewidth}{@{\extracolsep{\fill}}lrrr@{}}
\hline
\multicolumn{4}{@{}l}{\textit{RAID}} \\
\hline
Split & Total & Human & Machine \\
\hline
Train & 303,247 & 8,571 & 294,676 \\
Valid & 33,694 & 952 & 32,742 \\
Test & 112,317 & 3,232 & 109,085 \\
\hline
\end{tabular*}
\end{minipage}
\caption{
Split sizes after preprocessing.
Each panel reports the total number of examples and the counts for each class for one evaluation setting.
MAGE CDCM denotes the cross-domain cross-model setting of MAGE.
For M4, Valid corresponds to the official development split and is not merged into training.
For RAID, Train and Valid are obtained by a deterministic 90\%/10\% split of the processed RAID training data.
}
\label{tab:dataset_split_sizes}
\end{table*}

\begin{table*}[t!]
\centering
\small
\renewcommand{\arraystretch}{1.08}
\setlength{\tabcolsep}{4.5pt}
\begin{minipage}[t]{0.48\textwidth}
\centering
\begin{tabular}{@{}lrrr@{}}
\hline
\multicolumn{4}{@{}l}{\textit{MAGE Unseen Domains}} \\
\hline
Domain & Total & Human & Machine \\
\hline
CMV & 4,917 & 2,403 & 2,514 \\
ELI5 & 6,351 & 3,156 & 3,195 \\
HellaSwag & 6,347 & 3,292 & 3,055 \\
ROC & 6,462 & 3,275 & 3,187 \\
SciGen & 4,789 & 2,538 & 2,251 \\
SQuAD & 5,004 & 2,508 & 2,496 \\
TLDR & 4,977 & 2,535 & 2,442 \\
WP & 6,236 & 3,099 & 3,137 \\
XSum & 6,537 & 3,283 & 3,254 \\
Yelp & 5,199 & 2,652 & 2,547 \\
\hline
\end{tabular}
\end{minipage}
\hfill
\begin{minipage}[t]{0.48\textwidth}
\centering
\begin{tabular}{@{}lrrr@{}}
\hline
\multicolumn{4}{@{}l}{\textit{MAGE Unseen Models}} \\
\hline
Generator Model Family & Total & Human & Machine \\
\hline
GLM-130B & 1,838 & 919 & 919 \\
LLaMA & 7,420 & 3,710 & 3,710 \\
BigScience & 5,386 & 2,693 & 2,693 \\
FLAN-T5 & 9,320 & 4,660 & 4,660 \\
OpenAI & 13,284 & 6,642 & 6,642 \\
EleutherAI & 2,884 & 1,442 & 1,442 \\
OPT & 16,024 & 8,012 & 8,012 \\
\hline
\end{tabular}
\end{minipage}
\caption{
Test sizes for the MAGE leave-one-out evaluations.
The left panel lists leave-one-domain-out scenarios, and the right panel lists leave-one-generator-family-out scenarios.
Each row corresponds to an independently trained leave-one-out evaluation scenario.
}
\label{tab:mage_leaveoneout_sizes}
\end{table*}

\paragraph{MAGE Splits.}
MAGE CDCM uses the cross-domain cross-model setting of MAGE \citep{li-etal-2024-mage}.
The processed split contains 10 source domains: CMV, ELI5, HellaSwag, ROC, SciGen, SQuAD, TLDR, WP, XSum, and Yelp.
Machine-generated texts are produced by 27 model variants grouped into seven Generator Model Families: LLaMA, BigScience, FLAN-T5, GLM-130B, EleutherAI, OpenAI, and OPT.
For MAGE Unseen Domains, each leave-one-domain-out scenario excludes one source domain from training and validation and evaluates on that domain.
For MAGE Unseen Models, each leave-one-generator-family-out scenario excludes one Generator Model Family from training and validation and evaluates on that family.
Specifically, we used LLaMA-7B, BLOOM-7B, FLAN-T5-Small, GLM-130B, GPT-J, GPT-3.5-Turbo and OPT-125M model for the unseen-model setting evaluation.
The test sizes for the leave-one-out scenarios are listed in Table~\ref{tab:mage_leaveoneout_sizes}.

\paragraph{M4 Split.}
M4 is a multi-generator, multi-domain, and multilingual benchmark for black-box machine-generated text detection \citep{wang2024m4}.
We use the SemEval-2024 M4 Subtask A multilingual protocol with the multilingual train, development, and test splits.
The development split is used to select the checkpoint and threshold and is not merged into training.
The training split spans nine sources, while the multilingual test split contains Arabic, German, Italian, and English.
The test-time generator labels include BLOOMZ, ChatGPT, Cohere, Davinci, Dolly, JAIS-30B, and LLaMA2 fine-tuned.

\paragraph{RAID Split.}
RAID is designed to evaluate detector robustness across domains, generators, decoding strategies, and adversarial perturbations \citep{dugan2024raid}.
In our experiments, we use the RAID split preprocessed by \citet{zeng2025human-ood}.
We apply a 90\%/10\% split to the processed RAID training data, stratified by binary label.
The resulting train split is used for fitting, valid is used to select the checkpoint and threshold, and the processed RAID test split is used for final evaluation.

RAID covers eight domains: abstracts, books, news, poetry, recipes, Reddit, reviews, and Wiki.
The machine-generated side spans several generator families, including GPT-series, Cohere, LLaMA, Mistral, and MPT variants, alongside human-written text.
The train and validation splits contain clean examples, while the final test split contains both clean and perturbed examples.
The perturbations include paraphrase, homoglyph, whitespace, zero-width-space, synonym, and spelling or formatting perturbations.
Thus, RAID evaluates test-time robustness to adversarial and decoding-related variation, while model selection is performed only on clean validation data.

\paragraph{MAGE Leave-one-out Settings.}
For MAGE Unseen Domains and MAGE Unseen Models, each leave-one-out scenario trains a separate detector.
Aggregate metrics are computed as arithmetic means over scenario-level results:
\[
m_{\mathrm{UD}}
=
\frac{1}{10}
\sum_{d \in \mathcal D_{\mathrm{UD}}}
m_d,
\qquad
m_{\mathrm{UM}}
=
\frac{1}{7}
\sum_{g \in \mathcal G_{\mathrm{UM}}}
m_g.
\]
Thus, MAGE leave-one-out results are scenario-wise macro averages.
All metrics in Appendix~\ref{app:full_results} are reported as percentages.
\section{Ablation Definitions and Additional Analyses}
\label{app:ablation_details}

This appendix provides the exact definitions of the nonstandard variants evaluated in Table~\ref{tab:ablation_results} and reports the weighted-inference and mini-batch-hub analyses.
The ablation results and their interpretation are presented in Section~\ref{sec:deepfake_ablation} and are not repeated here.

\subsection{Ablation Definitions}
\label{app:ablation_definitions}

\paragraph{Prototype Repulsion.}
We replace Prototype Positioning with a direct same-class prototype repulsion loss.
Because the prototypes are normalized, their inner products correspond to cosine similarity:
\begin{equation}
\begin{aligned}
\mathcal L_{\mathrm{rep}}
&=
\frac{1}{2R(R-1)}
\sum_{c=0}^{1}
\sum_{r=1}^{R}
\sum_{k\neq r}
\left[
\max\left(0,p_{c,r}^{\top}p_{c,k}\right)
\right]^2 .
\end{aligned}
\end{equation}
This objective separates same-class prototype vectors without using sample-dependent positioning targets.
All other objectives and training settings remain unchanged.

\paragraph{Raw-Space Prototype Positioning.}
We also evaluate Prototype Positioning without removing the class-hub direction.
For each prototype, the assignment-weighted aggregate is formed directly from the normalized sample embeddings:
\begin{equation}
g_{c,r}
=
\sum_{i\in\mathcal B_c}q_{i,r}z_i,
\qquad
\bar z_{c,r}
=
\operatorname{norm}_{\varepsilon}(g_{c,r}).
\end{equation}
The corresponding objective is
\begin{equation}
\mathcal L_{\mathrm{PP\text{-}raw}}
=
-\frac{1}{2R}
\sum_{c=0}^{1}
\sum_{r=1}^{R}
\log
\frac{
\phi(\bar z_{c,r},p_{c,r})
}{
\sum_{p\in\mathcal P}\phi(\bar z_{c,r},p)
}.
\end{equation}
This variant retains prototype-specific sample aggregation while omitting the decomposition into class-shared and residual components.

\paragraph{Encoder Backbones.}
The encoder rows in Table~\ref{tab:ablation_results} cover backbones from the SimCSE, RoBERTa, BERT, E5, and BGE families
\citep{gao-etal-2021-simcse,liu2019roberta,devlin-etal-2019-bert,wang2022text,xiao2024cpack}.
All backbone variants use the same prototype configuration and training protocol.

\subsection{Weighted Prototype Inference}
\label{app:weighted_inference}

The main method scores each class using its maximum prototype similarity.
We additionally evaluate a weighted class score using the same temperature as the within-class assignments in Equation~\ref{eq:assignment}:
\begin{align}
w_{c,r}(z)
&=
\frac{
\exp(z^\top p_{c,r}/\tau)
}{
\sum_{k=1}^{R}\exp(z^\top p_{c,k}/\tau)
},
\\
\widetilde{s}_c(z)
&=
\sum_{r=1}^{R}
w_{c,r}(z)\,
z^\top p_{c,r},
\\
\widetilde{S}(z)
&=
\widetilde{s}_1(z)-\widetilde{s}_0(z).
\end{align}
All model parameters remain frozen; only the rule used to combine prototype similarities within each class is changed.

\begin{table}[t]
\centering
\small
\renewcommand{\arraystretch}{1.08}
\begin{tabular}{lcc}
\hline
Setting & Hard max & Weighted \\
\hline
MAGE CDCM & 95.14 & 95.08 \\
RAID      & 88.18 & 88.25 \\
M4        & 86.03 & 88.49 \\
\hline
\end{tabular}
\caption{AvgRec under hard-maximum and weighted prototype inference.}
\label{tab:weighted-inference}
\end{table}

Weighted inference leaves MAGE CDCM and RAID nearly unchanged and improves M4 without retraining.
Its effect is therefore most pronounced under the M4 evaluation setting.

\subsection{Mini-Batch Hub Stability}
\label{app:hub_stability}

We sample frozen training embeddings using the same sampler and batch size as training and compare each mini-batch hub with the corresponding class direction computed from the full training set.

\begin{table}[t]
\centering
\scriptsize
\renewcommand{\arraystretch}{1.08}
\setlength{\tabcolsep}{3.5pt}
\begin{tabular}{lcc}
\hline
Dataset & M/H per batch & Hub cosine M/H \\
\hline
MAGE CDCM & 181 / 75  & 0.9995 / 0.9963 \\
RAID      & 249 / 7   & 1.0000 / 0.9883 \\
M4        & 132 / 124 & 0.9996 / 0.9994 \\
\hline
\end{tabular}
\caption{Mean mini-batch composition and cosine similarity between mini-batch hubs and the corresponding full-training-data class directions. Batch counts are rounded to the nearest sample.}
\label{tab:hub-stability}
\end{table}

The mini-batch hubs remain closely aligned with the corresponding full-data directions across all three datasets.
This alignment is also maintained for the human class in RAID, despite its substantially smaller batch count.

\subsection{Seed Variation of Prototype Positioning}
\label{app:seed_variation}

We repeat the comparison between the full model and the variant without $\mathcal{L}_{\mathrm{PP}}$ on MAGE CDCM with three seeds.
For each seed, the two runs share the K-Means initialization, the initial model state, and the mini-batch order, on an NVIDIA RTX 3090 GPU.
Across the three seeds, AvgRec is 95.00 $\pm$ 0.08 with Prototype Positioning and 94.69 $\pm$ 0.19 without it (mean $\pm$ SD), AUROC is 98.45 $\pm$ 0.35 and 97.95 $\pm$ 0.15, and FPR95 is 5.06 $\pm$ 0.18 and 5.66 $\pm$ 0.33, respectively.
Prototype Positioning gives higher AvgRec and AUROC and lower FPR95 in all three pairs, with AvgRec gains ranging from 0.04 to 0.58 points.

\subsection{Comparison with SAMP}
\label{app:samp}

SAMP \citep{xu2026samp} also represents each class with multiple prototypes but uses generator labels as additional supervision.
We reimplement it in our training pipeline with eight prototypes per class.
Generator labels are used only in a source-contrastive loss among machine-generated texts and not at inference.
The objective combines a classification loss on the maximum prototype logit (scale 0.1), a balanced-assignment loss with Sinkhorn assignments ($\varepsilon=0.1$, 100 iterations, $\tau=0.07$) weighted by 0.5, and the source-contrastive loss ($\tau=0.07$).
Prototypes are updated by an exponential moving average with momentum 0.99, and inference uses the hard-maximum score with the validation-selected threshold.

Both detectors use seed 45 and share the initial model state, the K-Means prototype initialization, and the mini-batch order, on the same GPU.

\begin{table}[t]
\centering
\scriptsize
\renewcommand{\arraystretch}{1.08}
\setlength{\tabcolsep}{3.0pt}
\begin{tabular}{lccccc}
\hline
Method & AvgRec & HumanRec & MachineRec & AUROC & FPR95 \\
\hline
SAMP (reimpl.) & 93.38 & 92.92 & 93.85 & 97.62 & 9.49 \\
MD-ProTector & \textbf{95.01} & \textbf{95.46} & \textbf{94.55} & \textbf{98.65} & \textbf{5.03} \\
\hline
\end{tabular}
\caption{Comparison with our SAMP reimplementation on MAGE CDCM under matched initialization and mini-batch order (seed 45).}
\label{tab:samp}
\end{table}

MD-ProTector outperforms the SAMP reimplementation on all five metrics without using generator labels.

\section{Details of Prototype Analysis}
\label{app:prototype-analysis-details}
\raggedbottom

This appendix reports the residual-to-full preference analysis, defines the writing cues used in Section~\ref{sec:prototype-analysis}, details the writing-cue prediction and within-domain comparisons, and provides complete summaries of assigned texts together with their split-half stability.
The cue definitions are fixed before prototype-wise aggregation.

\subsection{Residual-to-Full Prototype Preference}
\label{app:prototype-preference-agreement}

For a class-$c$ vector, define its normalized hub-removed representation as
\begin{equation}
R_c(v)=\mathrm{norm}\left(v-(v^\top h_c)h_c\right).
\end{equation}
For each test sample $i$, we compare the same-class full-space and hub-removed preferences,
\begin{align}
r_i^{\mathrm{full}}
&=\arg\max_r z_i^\top p_{y_i,r},\\
r_i^{\perp}
&=\arg\max_r R_{y_i}(z_i)^\top R_{y_i}(p_{y_i,r}).
\end{align}
The analysis uses the same $R=8$ hard-max checkpoint and 7,200-sample MAGE CDCM test capture, containing 3,600 machine and 3,600 human samples.
The two preferences agree for 98.9\% of machine samples and 98.3\% of human samples.
Thus, the prototype preferred after removing the class hub is almost always the same prototype selected in the full space at inference.

\subsection{Writing Cue Definitions}
\label{app:prototype-features}

We compute 21 post-hoc measurements for every MAGE CDCM test sample.
Table~\ref{tab:prototype-feature-inventory} defines the cue names used in the main paper, and Tables~\ref{tab:machine-prototype-features} and~\ref{tab:human-prototype-features} report the complete prototype summaries.
The analysis implementation fixes the lexicons, regular expressions, tokenizer, and sentence-segmentation rules before prototype-wise aggregation.

\begin{table*}[t]
\centering
\footnotesize
\renewcommand{\arraystretch}{1.08}
\setlength{\tabcolsep}{4.0pt}
\begin{tabular*}{\textwidth}{@{\extracolsep{\fill}}p{0.14\textwidth}p{0.32\textwidth}p{0.14\textwidth}p{0.32\textwidth}@{}}
\hline
Writing cue & Measured as & Writing cue & Measured as \\
\hline
Words & Number of word tokens
& Instruction & Imperative or procedural expressions \\
Sentences & Number of detected sentences
& Attribution & Reporting or attribution expressions \\
Sentence length & Word count divided by sentence count
& Headings & Section-heading patterns \\
Lexical diversity & Unique word types divided by word count
& Lists & List bullets or enumerated items \\
Token repetition & Fraction of word tokens occurring more than once
& Templates & Bracketed template or pipeline markers \\
Single-use vocabulary & Fraction of word types occurring once
& Academic & Academic-register expressions \\
Word length & Mean characters per word
& Review vocabulary & Any whole-word match to \textit{star(s), service, staff, food, price, hotel, restaurant, recommend, delicious, favorite,} or \textit{ordered} \\
Punctuation & Punctuation characters divided by text length
& First person & Any of \textit{I, me, my, mine, we, us, our, ours} (case-insensitive) \\
Phrase repetition & Repeated word trigrams divided by available trigrams
& Second person & Any of \textit{you, your, yours} (case-insensitive) \\
Questions & Question-form markers
& Links/markup & URLs or markup patterns \\
Quotation & Quotation or quoted-speech markers
& & \\
\hline
\end{tabular*}
\caption{The 21 post-hoc measurements used in the prototype analysis. Continuous cues use word and sentence statistics. Binary cues use fixed lexicons or regular expressions.}
\label{tab:prototype-feature-inventory}
\end{table*}
\FloatBarrier
The review-vocabulary cue uses the fixed, case-insensitive term list shown in Table~\ref{tab:prototype-feature-inventory}; it is not a learned review classifier.

\subsection{Writing-Cue Prediction from Prototype Assignments}
\label{app:cue_prediction}

For each of the 21 writing cues in Table~\ref{tab:prototype-feature-inventory}, we fit a regression from domain indicators to the cue and then add each text's prototype assignment within its true class.
We report the relative reduction in held-out prediction error from adding the assignment.
The models with and without Prototype Positioning are trained with the same prototype count and the remaining objectives fixed, so both are scored alike.

For the models reported in the main text, the assignment reduces error by 20.16\% for machine texts and 5.90\% for human texts with PP, compared with 0.58\% and 0.18\% without PP.
PP lowers the error on 19 of 21 cues for machine texts and 18 of 21 cues for human texts, relative to the model without PP.
K-Means clusters of frozen embeddings reduce the error by 11.74--13.49\% for machine texts and 0.51--0.67\% for human texts.

Across two additional training runs per model, the reductions are 16.63--19.81\% with PP against 0.34--0.45\% without PP for machine texts, and 4.67--4.86\% against 0.62--1.76\% for human texts.
In these runs, PP lowers the error on 21 and 20 machine cues, respectively, and on 16 human cues in both runs.

\subsection{Within-Domain Cue Differences}
\label{app:within_domain}

For each prototype, we compared the texts assigned to it with the remaining texts of the same class within each MAGE domain that contained at least 20 texts on both sides, and averaged these within-domain differences with weights proportional to the number of texts of that class in each domain.
Binary cues are reported as differences in occurrence rates (percentage points) and continuous cues in their original units.
The 95\% intervals are percentile intervals from 2,000 bootstrap resamples of the texts within each class, and two-sided p-values from a normal approximation to the weighted difference were adjusted with the Benjamini--Hochberg procedure across all 336 prototype--cue comparisons.
Both procedures treat texts as independent, because the generation sources that would link related MAGE texts are not available.

Of the 336 comparisons for the full model, 195 are significant after adjustment (adjusted $p<0.05$), and 30 are undefined because no domain meets the inclusion criterion or the variance is zero.
Table~\ref{tab:within-domain} reports, for each prototype, the significant cue with the largest absolute difference after dividing by the class standard deviation of that cue.
H0--H3 each have only one qualifying domain, so their differences are comparisons within that domain.

\begin{table}[t]
\centering
\scriptsize
\renewcommand{\arraystretch}{1.08}
\setlength{\tabcolsep}{2.8pt}
\begin{tabular}{llrcc}
\hline
Proto. ($n$) & Cue & Difference & 95\% interval & Dom. \\
\hline
M0 (525) & Lexical diversity & $+0.21$ & [0.20, 0.22] & 8 \\
M1 (424) & Words & $+319.0$ & [294.6, 344.7] & 8 \\
M2 (541) & Token repetition & $+0.39$ & [0.38, 0.40] & 9 \\
M3 (479) & Words & $-215.1$ & [$-224.6$, $-205.2$] & 8 \\
M4 (459) & Words & $-154.2$ & [$-172.1$, $-136.7$] & 4 \\
M5 (386) & Lexical diversity & $+0.15$ & [0.13, 0.18] & 3 \\
M6 (475) & Words & $-91.9$ & [$-105.3$, $-77.5$] & 6 \\
M7 (311) & Academic & $+18.19$ pp & [9.52, 27.20] & 3 \\
\hline
H0 (147) & Review vocabulary & $+23.22$ pp & [14.31, 31.71] & 1 \\
H1 (235) & Quotation & $+55.03$ pp & [46.62, 63.76] & 1 \\
H2 (242) & -- & -- & -- & -- \\
H3 (71) & Token repetition & $-0.18$ & [$-0.21$, $-0.16$] & 1 \\
H4 (583) & Words & $+389.3$ & [335.5, 450.4] & 4 \\
H5 (1370) & Questions & $-14.95$ pp & [$-18.44$, $-11.40$] & 8 \\
H6 (510) & Quotation & $-14.24$ pp & [$-18.75$, $-9.38$] & 4 \\
H7 (442) & Sentences & $-16.5$ & [$-18.1$, $-15.0$] & 4 \\
\hline
\end{tabular}
\caption{Largest standardized within-domain cue difference for each prototype relative to the remaining texts of its class. Units follow Table~\ref{tab:prototype-feature-inventory}; ``Dom.'' is the number of qualifying domains. H2 has no significant cue.}
\label{tab:within-domain}
\end{table}

We also tested the cue descriptions in Tables~\ref{tab:machine-prototype-features} and~\ref{tab:human-prototype-features} within domains.
M1 uses quotation 44.62 pp more often than the remaining machine texts ([39.81, 49.33], eight domains), and the descriptions of M1 (instruction), M3, M5, M7, H0, and H4 likewise remain significant.
The template description of M4 ($-0.24$ pp, adjusted $p=0.13$), the links description of H1 ($+6.22$ pp, adjusted $p=0.086$, one domain), and the academic description of H6 ($+4.88$ pp, adjusted $p=0.051$) keep their direction but are not significant.
H7 uses first-person expressions no more often than other human texts in the same domains ($-0.62$ pp, adjusted $p=0.83$), and the template markers of H5 occur only in HellaSwag, whose human texts are almost all assigned to H5, so these two descriptions reflect domain composition.

\subsection{Stability of Prototype Descriptions}
\label{app:prototype-feature-stability}

We recompute the 21-dimensional feature-association vector for every prototype over 50 label-by-domain-stratified split halves.
Within each half, each feature is standardized across the 16 prototypes before comparing the two vectors for the same prototype.
The mean split-half cosine is 0.837 for machine prototypes and 0.908 for human prototypes.
All 16 prototypes receive at least 20 hard assignments in the 7,200-sample test capture.

\subsection{Complete Prototype Summaries}
Tables~\ref{tab:machine-prototype-features} and~\ref{tab:human-prototype-features} summarize the texts assigned to all machine and human prototypes. The cue names match the inventory in Table~\ref{tab:prototype-feature-inventory}.

\begin{table*}[t]
\centering
\footnotesize
\renewcommand{\arraystretch}{1.10}
\setlength{\tabcolsep}{3.0pt}
\begin{tabular*}{\textwidth}{@{\extracolsep{\fill}}p{0.10\textwidth}p{0.14\textwidth}p{0.17\textwidth}p{0.50\textwidth}@{}}
\hline
Prototype ($n$) & Top domain & Top generator & Distinctive characteristics \\
\hline
M0 (525) & XSum 34.7\% & OPT-30B 11.6\% & - \\
M1 (424) & SQuAD 19.8\% & OPT-6.7B 11.6\% & Quotation occurs in 77.6\% and instruction expressions in 78.1\% of assigned texts. \\
M2 (541) & TLDR 14.8\% & LLaMA-13B 17.4\% & - \\
M3 (479) & HellaSwag 22.8\% & FLAN-T5-xl 15.0\% & Academic expressions occur in 0.2\% of assigned texts and 6.9\% of the remaining machine texts. \\
M4 (459) & ELI5 29.8\% & text-davinci-003 22.4\% & Template markers occur in 0.9\% of assigned texts and 6.1\% of the remaining machine texts. \\
M5 (386) & Yelp 62.7\% & BLOOM-7B 7.5\% & Review vocabulary occurs in 58.8\% of assigned texts. \\
M6 (475) & ROCStories 30.3\% & GPT-3.5-Turbo 24.8\% & - \\
M7 (311) & SciGen 74.6\% & GPT-3.5-Turbo 6.4\% & Academic expressions occur in 49.2\% of assigned texts. \\
\hline
\end{tabular*}
\caption{Complete summaries of texts assigned to the machine prototypes. Top domain and top generator denote the largest metadata shares within each assigned group. Distinctive characteristics report cue frequencies or comparisons with the remaining machine texts; within-domain tests are given in Appendix~\ref{app:within_domain}.}
\label{tab:machine-prototype-features}
\end{table*}

\begin{table*}[t]
\centering
\footnotesize
\renewcommand{\arraystretch}{1.10}
\setlength{\tabcolsep}{3.5pt}
\begin{tabular*}{\textwidth}{@{\extracolsep{\fill}}p{0.11\textwidth}p{0.20\textwidth}p{0.61\textwidth}@{}}
\hline
Prototype ($n$) & Top domain & Distinctive characteristics \\
\hline
H0 (147) & Yelp 93.2\% & First-person expressions occur in 98.6\% and review vocabulary in 85.7\% of assigned texts. \\
H1 (235) & ELI5 98.7\% & Links or markup occur in 14.0\% of assigned texts. \\
H2 (242) & SQuAD 97.9\% & - \\
H3 (71) & Yelp 95.8\% & - \\
H4 (583) & WritingPrompts 44.3\% & First-person expressions occur in 94.0\% of assigned texts. \\
H5 (1370) & ROCStories 30.3\% & Template markers occur in 15.0\% of assigned texts, all from HellaSwag. \\
H6 (510) & TLDR 41.4\%, SciGen 40.6\% & Academic expressions occur in 33.5\% of assigned texts. \\
H7 (442) & XSum 37.6\% & First-person expressions occur in 83.7\% of assigned texts, a rate similar to other human texts in the same domains. \\
\hline
\end{tabular*}
\caption{Complete summaries of texts assigned to the human prototypes. Top domain denotes the largest metadata share within each assigned group. Distinctive characteristics report cue frequencies or comparisons with the remaining human texts; within-domain tests are given in Appendix~\ref{app:within_domain}.}
\label{tab:human-prototype-features}
\end{table*}

\FloatBarrier

\section{Full Results}
\label{app:full_results}

We provide full results for all methods, including aggregate evaluation settings and detailed MAGE leave-one-out scenarios. 
Each table reports AvgRec, HumanRec, MachineRec, F1, Accuracy, AUROC, AUPR, and FPR95.
MAGE-UD and MAGE-UM denote MAGE Unseen Domains and MAGE Unseen Models, respectively.

\begin{table*}[p]
\centering
\scriptsize
\renewcommand{\arraystretch}{1.05}
\setlength{\tabcolsep}{2.7pt}
\begin{tabular*}{\textwidth}{@{\extracolsep{\fill}}llcccccccc@{}}
\hline
Setting & Scenario & AvgRec & HumanRec & MachineRec & F1 & Acc & AUROC & AUPR & FPR95$\downarrow$ \\
\hline
\multicolumn{10}{l}{\textit{Aggregate settings}} \\
MAGE CDCM & Mixed & 90.79 & 83.18 & 98.40 & 90.05 & 90.71 & 98.36 & 98.52 & 4.78 \\
M4 & Language shift & 76.68 & 54.56 & 98.80 & 70.00 & 77.67 & 95.80 & 95.55 & 18.63 \\
RAID & Adversarial/decoding & 86.81 & 74.60 & 99.02 & 71.80 & 98.31 & 89.48 & 77.32 & 93.90 \\
MAGE-UD & Average & 67.99 & 37.04 & 98.94 & 51.14 & 67.55 & 89.30 & 90.79 & 56.25 \\
MAGE-UM & Average & 89.78 & 85.85 & 93.71 & 89.41 & 89.78 & 97.03 & 96.93 & 12.37 \\
\hline
\multicolumn{10}{l}{\textit{MAGE Unseen Domains}} \\
UD & ROC & 53.78 & 7.85 & 99.72 & 14.52 & 53.16 & 88.75 & 88.25 & 46.38 \\
UD & HellaSwag & 66.87 & 37.24 & 96.50 & 53.02 & 65.76 & 91.04 & 89.00 & 28.25 \\
UD & XSum & 55.86 & 12.85 & 98.86 & 22.55 & 55.67 & 70.42 & 74.85 & 96.22 \\
UD & Yelp & 67.70 & 35.78 & 99.61 & 52.56 & 67.05 & 91.23 & 93.50 & 69.61 \\
UD & TLDR & 61.80 & 24.54 & 99.06 & 39.12 & 61.10 & 94.20 & 93.10 & 19.25 \\
UD & SciGen & 69.66 & 40.39 & 98.93 & 57.15 & 67.91 & 93.65 & 94.50 & 24.21 \\
UD & WP & 75.59 & 51.73 & 99.46 & 67.94 & 75.74 & 87.75 & 92.07 & 91.81 \\
UD & CMV & 79.57 & 60.13 & 99.01 & 74.62 & 80.01 & 91.82 & 94.42 & 81.50 \\
UD & SQuAD & 67.68 & 35.49 & 99.88 & 52.34 & 67.61 & 89.16 & 92.44 & 83.20 \\
UD & ELI5 & 81.43 & 64.45 & 98.40 & 77.62 & 81.53 & 94.94 & 95.79 & 22.10 \\
\hline
\multicolumn{10}{l}{\textit{MAGE Unseen Models}} \\
UM & LLaMA & 91.33 & 85.74 & 96.93 & 90.82 & 91.33 & 98.17 & 97.84 & 7.12 \\
UM & FLAN-T5 & 85.04 & 83.61 & 86.48 & 84.82 & 85.04 & 92.67 & 92.62 & 33.43 \\
UM & EleutherAI & 92.02 & 84.26 & 99.79 & 91.35 & 92.02 & 99.37 & 99.57 & 0.76 \\
UM & GLM-130B & 91.62 & 86.07 & 97.17 & 91.13 & 91.62 & 98.25 & 98.32 & 6.31 \\
UM & OPT & 90.88 & 86.50 & 95.26 & 90.46 & 90.88 & 97.60 & 97.31 & 9.67 \\
UM & BigScience & 89.77 & 82.14 & 97.40 & 88.92 & 89.77 & 97.84 & 97.88 & 8.76 \\
UM & OpenAI & 87.78 & 92.65 & 82.91 & 88.35 & 87.78 & 95.28 & 95.00 & 20.52 \\
\hline
\end{tabular*}
\caption{
Full results for Binary CE.
}
\label{tab:full_binary_ce}
\end{table*}

\begin{table*}[p]
\centering
\scriptsize
\renewcommand{\arraystretch}{1.05}
\setlength{\tabcolsep}{2.7pt}
\begin{tabular*}{\textwidth}{@{\extracolsep{\fill}}llcccccccc@{}}
\hline
Setting & Scenario & AvgRec & HumanRec & MachineRec & F1 & Acc & AUROC & AUPR & FPR95$\downarrow$ \\
\hline
\multicolumn{10}{l}{\textit{Aggregate settings}} \\
MAGE CDCM & Mixed & 94.77 & 92.98 & 96.56 & 94.71 & 94.75 & 95.06 & 96.79 & 31.27 \\
M4 & Language shift & 83.84 & 72.37 & 95.30 & 81.54 & 84.35 & 89.23 & 92.21 & 72.72 \\
RAID & Adversarial/decoding & 77.67 & 58.94 & 96.39 & 42.00 & 95.32 & 78.04 & 45.77 & 88.26 \\
MAGE-UD & Average & 75.11 & 56.82 & 93.41 & 67.01 & 74.84 & 75.15 & 83.68 & 84.16 \\
MAGE-UM & Average & 90.92 & 93.49 & 88.34 & 91.40 & 90.92 & 91.89 & 93.90 & 32.59 \\
\hline
\multicolumn{10}{l}{\textit{MAGE Unseen Domains}} \\
UD & ROC & 55.72 & 12.06 & 99.37 & 21.41 & 55.12 & 55.71 & 75.80 & 94.35 \\
UD & HellaSwag & 72.64 & 52.58 & 92.70 & 65.99 & 71.89 & 72.77 & 82.82 & 90.23 \\
UD & XSum & 65.99 & 35.58 & 96.40 & 51.14 & 65.86 & 66.00 & 79.24 & 92.52 \\
UD & Yelp & 50.08 & 37.22 & 62.94 & 43.07 & 49.82 & 49.15 & 51.03 & 95.43 \\
UD & TLDR & 72.57 & 47.14 & 97.99 & 63.24 & 72.09 & 72.68 & 85.15 & 90.73 \\
UD & SciGen & 84.35 & 72.66 & 96.05 & 82.49 & 83.65 & 84.52 & 91.29 & 82.44 \\
UD & WP & 92.04 & 85.87 & 98.21 & 91.51 & 92.08 & 92.13 & 95.49 & 65.25 \\
UD & CMV & 91.64 & 86.27 & 97.02 & 91.10 & 91.76 & 91.98 & 95.07 & 64.68 \\
UD & SQuAD & 79.92 & 60.89 & 98.96 & 75.20 & 79.88 & 79.99 & 89.49 & 87.35 \\
UD & ELI5 & 86.17 & 77.92 & 94.43 & 84.90 & 86.22 & 86.53 & 91.35 & 78.62 \\
\hline
\multicolumn{10}{l}{\textit{MAGE Unseen Models}} \\
UM & LLaMA & 93.48 & 93.29 & 93.67 & 93.46 & 93.48 & 94.18 & 95.78 & 30.22 \\
UM & FLAN-T5 & 80.98 & 94.25 & 67.70 & 83.21 & 80.98 & 85.20 & 87.79 & 41.14 \\
UM & EleutherAI & 96.05 & 92.30 & 99.79 & 95.89 & 96.05 & 96.08 & 97.99 & 35.18 \\
UM & GLM-130B & 94.29 & 92.38 & 96.19 & 94.18 & 94.29 & 94.51 & 96.31 & 36.86 \\
UM & OPT & 93.26 & 93.58 & 92.94 & 93.28 & 93.26 & 93.82 & 95.35 & 27.57 \\
UM & BigScience & 93.15 & 92.94 & 93.35 & 93.13 & 93.15 & 93.39 & 95.07 & 33.84 \\
UM & OpenAI & 85.21 & 95.69 & 74.72 & 86.61 & 85.21 & 86.06 & 88.98 & 23.31 \\
\hline
\end{tabular*}
\caption{
Full results for SupCon.
}
\label{tab:full_supcon}
\end{table*}

\begin{table*}[p]
\centering
\scriptsize
\renewcommand{\arraystretch}{1.05}
\setlength{\tabcolsep}{2.7pt}
\begin{tabular*}{\textwidth}{@{\extracolsep{\fill}}llcccccccc@{}}
\hline
Setting & Scenario & AvgRec & HumanRec & MachineRec & F1 & Acc & AUROC & AUPR & FPR95$\downarrow$ \\
\hline
\multicolumn{10}{l}{\textit{Aggregate settings}} \\
MAGE CDCM & Mixed & 94.84 & 91.87 & 97.81 & 94.71 & 94.80 & 95.81 & 97.36 & 23.07 \\
M4 & Language shift & 92.74 & 87.92 & 97.57 & 92.27 & 92.96 & 92.74 & 95.38 & 59.62 \\
RAID & Adversarial/decoding & 87.68 & 80.17 & 95.20 & 46.85 & 94.77 & 91.08 & 62.01 & 67.70 \\
MAGE-UD & Average & 76.72 & 55.75 & 97.69 & 67.09 & 76.42 & 79.49 & 88.39 & 82.17 \\
MAGE-UM & Average & 91.69 & 92.08 & 91.30 & 91.85 & 91.69 & 92.87 & 94.61 & 27.14 \\
\hline
\multicolumn{10}{l}{\textit{MAGE Unseen Domains}} \\
UD & ROC & 53.43 & 7.08 & 99.78 & 13.20 & 52.80 & 54.32 & 75.86 & 94.53 \\
UD & HellaSwag & 78.62 & 66.01 & 91.23 & 75.81 & 78.15 & 82.84 & 87.35 & 81.69 \\
UD & XSum & 61.16 & 24.03 & 98.28 & 38.23 & 60.99 & 64.78 & 78.90 & 92.83 \\
UD & Yelp & 77.81 & 56.30 & 99.33 & 71.74 & 77.38 & 80.09 & 89.81 & 87.32 \\
UD & TLDR & 70.70 & 42.17 & 99.22 & 59.01 & 70.16 & 73.52 & 86.22 & 90.49 \\
UD & SciGen & 80.49 & 63.83 & 97.16 & 76.74 & 79.49 & 83.38 & 90.87 & 83.99 \\
UD & WP & 88.55 & 78.28 & 98.82 & 87.23 & 88.61 & 91.12 & 95.10 & 70.38 \\
UD & CMV & 90.66 & 82.40 & 98.93 & 89.80 & 90.85 & 92.74 & 95.86 & 63.16 \\
UD & SQuAD & 78.77 & 58.33 & 99.20 & 73.31 & 78.72 & 82.38 & 90.72 & 85.63 \\
UD & ELI5 & 86.99 & 79.02 & 94.96 & 85.84 & 87.04 & 89.74 & 93.26 & 71.69 \\
\hline
\multicolumn{10}{l}{\textit{MAGE Unseen Models}} \\
UM & LLaMA & 93.77 & 93.58 & 93.96 & 93.76 & 93.77 & 95.41 & 96.27 & 6.73 \\
UM & FLAN-T5 & 86.63 & 90.88 & 82.38 & 87.18 & 86.63 & 87.98 & 90.19 & 41.95 \\
UM & EleutherAI & 95.46 & 90.98 & 99.93 & 95.25 & 95.46 & 96.46 & 98.21 & 28.71 \\
UM & GLM-130B & 93.58 & 90.32 & 96.84 & 93.36 & 93.58 & 94.63 & 96.43 & 38.17 \\
UM & OPT & 93.32 & 93.00 & 93.65 & 93.30 & 93.32 & 94.24 & 95.53 & 16.50 \\
UM & BigScience & 93.32 & 90.61 & 96.03 & 93.13 & 93.32 & 94.38 & 96.07 & 35.10 \\
UM & OpenAI & 85.77 & 95.21 & 76.33 & 87.00 & 85.77 & 87.01 & 89.55 & 22.82 \\
\hline
\end{tabular*}
\caption{
Full results for DeTeCtive.
}
\label{tab:full_detective}
\end{table*}

\begin{table*}[p]
\centering
\scriptsize
\renewcommand{\arraystretch}{1.05}
\setlength{\tabcolsep}{2.7pt}
\begin{tabular*}{\textwidth}{@{\extracolsep{\fill}}llcccccccc@{}}
\hline
Setting & Scenario & AvgRec & HumanRec & MachineRec & F1 & Acc & AUROC & AUPR & FPR95$\downarrow$ \\
\hline
\multicolumn{10}{l}{\textit{Aggregate settings}} \\
MAGE CDCM & Mixed & 94.43 & 95.18 & 93.67 & 94.54 & 94.44 & 98.47 & 98.56 & 6.15 \\
M4 & Language shift & 81.08 & 63.13 & 99.03 & 76.90 & 81.89 & 95.20 & 96.32 & 12.53 \\
RAID & Adversarial/decoding & 86.17 & 76.36 & 95.97 & 48.89 & 95.41 & 87.47 & 51.98 & 96.52 \\
MAGE-UD & Average & 79.08 & 63.46 & 94.71 & 72.57 & 78.90 & 92.11 & 92.02 & 33.25 \\
MAGE-UM & Average & 90.71 & 95.22 & 86.19 & 91.38 & 90.71 & 96.71 & 96.79 & 13.33 \\
\hline
\multicolumn{10}{l}{\textit{MAGE Unseen Domains}} \\
UD & ROC & 59.50 & 20.06 & 98.93 & 33.13 & 58.96 & 87.08 & 86.51 & 45.09 \\
UD & HellaSwag & 83.92 & 80.01 & 87.82 & 83.65 & 83.77 & 92.36 & 90.10 & 23.54 \\
UD & XSum & 62.12 & 28.21 & 96.04 & 42.69 & 61.97 & 83.12 & 82.25 & 63.92 \\
UD & Yelp & 81.10 & 66.52 & 95.68 & 77.95 & 80.80 & 91.98 & 93.14 & 48.02 \\
UD & TLDR & 70.76 & 45.05 & 96.48 & 60.70 & 70.28 & 87.36 & 87.99 & 58.11 \\
UD & SciGen & 85.64 & 78.17 & 93.11 & 84.84 & 85.20 & 95.24 & 95.45 & 18.84 \\
UD & WP & 86.65 & 76.93 & 96.37 & 85.19 & 86.71 & 97.27 & 97.19 & 11.57 \\
UD & CMV & 89.23 & 84.14 & 94.31 & 88.53 & 89.34 & 96.60 & 96.71 & 16.91 \\
UD & SQuAD & 83.67 & 70.33 & 97.00 & 81.16 & 83.63 & 95.26 & 95.62 & 25.00 \\
UD & ELI5 & 88.27 & 85.17 & 91.36 & 87.84 & 88.29 & 94.86 & 95.24 & 21.56 \\
\hline
\multicolumn{10}{l}{\textit{MAGE Unseen Models}} \\
UM & LLaMA & 94.30 & 95.66 & 92.94 & 94.38 & 94.30 & 98.35 & 98.34 & 6.44 \\
UM & FLAN-T5 & 79.39 & 94.87 & 63.91 & 82.15 & 79.39 & 91.83 & 91.91 & 36.52 \\
UM & EleutherAI & 97.16 & 95.08 & 99.24 & 97.10 & 97.16 & 99.04 & 99.34 & 0.55 \\
UM & GLM-130B & 94.72 & 95.10 & 94.34 & 94.74 & 94.72 & 98.40 & 98.52 & 5.66 \\
UM & OPT & 91.20 & 94.22 & 88.18 & 91.46 & 91.20 & 96.92 & 97.01 & 13.27 \\
UM & BigScience & 92.65 & 94.65 & 90.64 & 92.79 & 92.65 & 97.44 & 97.55 & 9.84 \\
UM & OpenAI & 85.54 & 96.99 & 74.09 & 87.02 & 85.54 & 94.96 & 94.85 & 21.02 \\
\hline
\end{tabular*}
\caption{
Full results for DSVDD.
}
\label{tab:full_dsvdd}
\end{table*}

\begin{table*}[p]
\centering
\scriptsize
\renewcommand{\arraystretch}{1.05}
\setlength{\tabcolsep}{2.7pt}
\begin{tabular*}{\textwidth}{@{\extracolsep{\fill}}llcccccccc@{}}
\hline
Setting & Scenario & AvgRec & HumanRec & MachineRec & F1 & Acc & AUROC & AUPR & FPR95$\downarrow$ \\
\hline
\multicolumn{10}{l}{\textit{Aggregate settings}} \\
MAGE CDCM & Mixed & 95.14 & 95.81 & 94.47 & 95.23 & 95.15 & 98.41 & 98.37 & 4.89 \\
M4 & Language shift & 86.03 & 76.87 & 95.20 & 84.42 & 86.45 & 92.35 & 93.08 & 57.00 \\
RAID & Adversarial/decoding & 88.18 & 82.52 & 93.84 & 42.25 & 93.51 & 95.41 & 71.41 & 27.78 \\
MAGE-UD & Average & 78.59 & 61.46 & 95.72 & 71.13 & 78.34 & 86.90 & 89.10 & 59.41 \\
MAGE-UM & Average & 91.34 & 95.63 & 87.05 & 91.91 & 91.34 & 96.48 & 96.16 & 11.44 \\
\hline
\multicolumn{10}{l}{\textit{MAGE Unseen Domains}} \\
UD & ROC & 56.68 & 13.92 & 99.44 & 24.33 & 56.10 & 86.11 & 86.41 & 55.26 \\
UD & HellaSwag & 75.84 & 59.51 & 92.18 & 71.37 & 75.23 & 78.73 & 82.41 & 89.89 \\
UD & XSum & 61.29 & 25.68 & 96.90 & 39.89 & 61.13 & 68.73 & 72.74 & 92.58 \\
UD & Yelp & 83.29 & 71.68 & 94.90 & 81.19 & 83.05 & 88.71 & 91.76 & 78.03 \\
UD & TLDR & 75.06 & 53.89 & 96.23 & 68.42 & 74.66 & 81.49 & 86.25 & 87.85 \\
UD & SciGen & 83.82 & 72.89 & 94.76 & 82.11 & 83.17 & 95.89 & 95.21 & 12.67 \\
UD & WP & 91.29 & 85.25 & 97.32 & 90.71 & 91.32 & 96.67 & 97.32 & 16.67 \\
UD & CMV & 90.30 & 85.06 & 95.54 & 89.67 & 90.42 & 95.56 & 96.41 & 22.37 \\
UD & SQuAD & 80.18 & 62.76 & 97.60 & 76.00 & 80.14 & 85.27 & 89.54 & 80.55 \\
UD & ELI5 & 88.17 & 84.00 & 92.33 & 87.61 & 88.19 & 91.87 & 92.99 & 58.20 \\
\hline
\multicolumn{10}{l}{\textit{MAGE Unseen Models}} \\
UM & LLaMA & 94.20 & 96.01 & 92.40 & 94.31 & 94.20 & 98.36 & 97.77 & 6.45 \\
UM & FLAN-T5 & 83.12 & 96.12 & 70.13 & 85.06 & 83.12 & 93.13 & 91.96 & 26.07 \\
UM & EleutherAI & 96.78 & 94.24 & 99.31 & 96.69 & 96.78 & 99.07 & 99.32 & 0.96 \\
UM & GLM-130B & 94.23 & 95.32 & 93.14 & 94.29 & 94.23 & 96.47 & 96.52 & 6.42 \\
UM & OPT & 93.02 & 94.88 & 91.15 & 93.14 & 93.02 & 96.48 & 96.32 & 9.09 \\
UM & BigScience & 93.00 & 95.95 & 90.05 & 93.20 & 93.00 & 97.68 & 97.71 & 8.47 \\
UM & OpenAI & 85.05 & 96.90 & 73.20 & 86.63 & 85.05 & 94.19 & 93.50 & 22.63 \\
\hline
\end{tabular*}
\caption{
Full results for MD-ProTector.
}
\label{tab:full_mdprotector}
\end{table*}

\end{document}